\documentclass[compsoc,  onecolumn]{IEEEtran}

\usepackage[center]{caption}
\usepackage[dvipsnames]{xcolor}

\usepackage{cite,url}
\usepackage{amsmath}
\usepackage{amssymb}
\usepackage{mathrsfs}
\usepackage{nameref}
\usepackage[inline]{enumitem}
\usepackage{soul}
\usepackage[colorlinks=true,linkcolor=blue,anchorcolor=black,citecolor=blue,filecolor=black,menucolor=black,runcolor=black,urlcolor=black]{hyperref}

\usepackage{float}
\usepackage{verbatim} 
\usepackage{makecell} 

\usepackage{pifont}
\newcommand{\xmark}{\ding{55}}
\newcommand{\cmark}{\ding{51}}
\definecolor{cadmiumgreen}{rgb}{0.0, 0.42, 0.24} 
\usepackage{bm} 

\usepackage{xr}
\usepackage{todonotes}
\AtBeginDocument{%
  \providecommand\BibTeX{{%
    \normalfont B\kern-0.5em{\scshape i\kern-0.25em b}\kern-0.8em\TeX}}}

\author{Denis Kleyko, Dmitri A. Rachkovskij, Evgeny Osipov, and Abbas Rahimi\\ 
\thanks{We would like to thank three reviewers, the editors, and Pentti Kanerva for their insightful feedback as well as Linda Rudin for the careful proofreading that contributed to the final shape of the survey. 
The work of DK was supported by the European Union's Horizon 2020 Programme under the Marie Skłodowska-Curie Individual Fellowship Grant (839179). The work of DK  was also supported in part by AFOSR FA9550-19-1-0241 and Intel's THWAI program.
The work of DAR was supported in part by the National Academy of Sciences of Ukraine (grant no. 0120U000122, 0121U000016, and 0117U002286), the Ministry of Education and Science of Ukraine (grant no. 0121U000228 and 0122U000818), and the Swedish Foundation for Strategic Research (SSF, grant no. UKR22-0024).
\textit{Denis Kleyko and Dmitri A. Rachkovskij contributed equally to this work.}
}
\thanks{D. Kleyko is with the Redwood Center for Theoretical Neuroscience at the University of California, Berkeley, CA 94720, USA and also with the Intelligent Systems Lab at Research Institutes of Sweden, 16440 Kista, Sweden. \mbox{E-mail}: \mbox{denkle@berkeley.edu}
}
\thanks{D. A. Rachkovskij is with International Research and Training Center for Information Technologies and Systems, 03680 Kiev, Ukraine. 
\mbox{E-mail}: \mbox{dar@infrm.kiev.ua}
 }

\thanks{E. Osipov is with the Department of Computer  Science Electrical and Space Engineering, Lule\aa{} University of Technology, 97187 Lule\aa{}, Sweden. \mbox{E-mail}: \mbox{evgeny.osipov@ltu.se}
}
\thanks{A. Rahimi is with IBM Research, 8803 Zurich,  Switzerland. \mbox{E-mail}: \mbox{abr@zurich.ibm.com}
}
}

\begin{document}

\title{
A Survey on Hyperdimensional Computing \\ aka Vector Symbolic Architectures, Part II: Applications, Cognitive Models, and Challenges
}

\maketitle

\begin{abstract}
This is Part~II of the two-part comprehensive survey devoted to a computing framework most commonly known under the names Hyperdimensional Computing and Vector Symbolic Architectures (HDC/VSA). Both  names refer to a  family of computational models that use high-dimensional distributed representations and rely on the algebraic properties of their key operations to incorporate the advantages of structured symbolic representations  and vector distributed representations. Holographic Reduced Representations~\cite{PlateHolographic1995, PlateHolographic2003} is an influential HDC/VSA model that is well-known in the machine learning domain and often used to refer to the whole family. 
However, for the sake of consistency, we use HDC/VSA to refer to the field.   

Part~I of this survey~\cite{KleykoSurveyVSA2021Part1} covered foundational aspects of the field, such as the historical context leading to the development of HDC/VSA, key elements of any HDC/VSA model, known HDC/VSA models, and the transformation of input data of various types into high-dimensional vectors suitable for HDC/VSA. This second part surveys existing applications, the role of HDC/VSA in cognitive computing and architectures, as well as directions for future work. Most of the applications lie within the Machine Learning/Artificial Intelligence domain, however, we also cover other applications to provide a complete picture.   
The survey is written to be useful for both newcomers and practitioners.
\end{abstract}


\newpage
\setcounter{tocdepth}{3}
\tableofcontents
\newpage

\section{Introduction}

This article is Part~II of the survey of a research field known under the names Hyperdimensional Computing, HDC (the term was introduced in~\cite{KanervaHyperdimensional2009}) and Vector Symbolic Architectures, VSA (the term was introduced in~\cite{GaylerJackendoff2003}).
As in Part I~\cite{KleykoSurveyVSA2021Part1}, below we will consistently use the joint name HDC/VSA when referring to the field. 
HDC/VSA is an umbrella term for a family of computational models that rely on mathematical properties of high-dimensional random spaces and use high-dimensional distributed representations called hypervectors (HVs) for a structured (``symbolic'') representation of data, while maintaining the advantages of traditional connectionist vector distributed representations.

First, let us briefly recapitulate the motivation for this survey. 
The main driving force behind the current interest in HDC/VSA is the
global trend of searching for computing paradigms alternative to the conventional (von Neumann) ones.
Examples of the new paradigms are  neuromorphic and nanoscalable  computing, where HDC/VSA is expected to play an important role (see~\cite{KleykoComputingParadigm2021} and references therein for perspective). 
Due to this surge of interest in HDC/VSA, the need for providing a broad overview of the field, which is currently missing, became  evident.
Therefore, this two-part survey extensively covers the state-of-the-art of the field in a form that is accessible to a wider audience.

\begin{table}[tb]
\tiny
\caption{A qualitative assessment of existing HDC/VSA literature that have some elements of survey. {\color{cadmiumgreen}\cmark} means that the article overviewed the area rather comprehensively, {\color{red} \xmark} means that the area was not covered at all while $\bm{\pm}$ indicates that the article partially addressed a particular topic, but either new results were reported since then or not all related work was covered.
}
\label{table:position:part2}
    \begin{center}
    \begin{tabular}{c c c c c c c c c c c c c c c c c c c} 
     \cline{2-19}
     &  \multicolumn{3}{|c|}{\makecell{ Deterministic \\
behavior}}  &  \multicolumn{3}{c|}{\makecell{ Similarity \\ estimation}} &  \multicolumn{3}{c|}{\makecell{ Classification}} &  \multicolumn{4}{c|}{\makecell{ Cognitive \\ computing}} &  \multicolumn{5}{c|}{\makecell{ Cognitive \\ architectures}} \\ \cline{2-19} 

& \multicolumn{1}{|c|}{\rotatebox[origin=c]{90}{\makecell{Automata \& instructions}}}
& \multicolumn{1}{c|}{\rotatebox[origin=c]{90}{\makecell{Data transmission}}}
& \multicolumn{1}{c|}{\rotatebox[origin=c]{90}{\makecell{String processing}}}
& \multicolumn{1}{c|}{\rotatebox[origin=c]{90}{\makecell{Context vectors}}}
& \multicolumn{1}{c|}{\rotatebox[origin=c]{90}{\makecell{Biomedical signals}}}
& \multicolumn{1}{c|}{\rotatebox[origin=c]{90}{\makecell{Images}}}
& \multicolumn{1}{c|}{\rotatebox[origin=c]{90}{\makecell{Feature vectors}}}
& \multicolumn{1}{c|}{\rotatebox[origin=c]{90}{\makecell{Images or their properties}}}
& \multicolumn{1}{c|}{\rotatebox[origin=c]{90}{\makecell{Structured data}}}
& \multicolumn{1}{c|}{\rotatebox[origin=c]{90}{\makecell{Holistic transformations}}}
& \multicolumn{1}{c|}{\rotatebox[origin=c]{90}{\makecell{Analogical reasoning}}}
& \multicolumn{1}{c|}{\rotatebox[origin=c]{90}{\makecell{Cognitive modeling}}}
& \multicolumn{1}{c|}{\rotatebox[origin=c]{90}{\makecell{Computer vision \\ and scene analysis}}}
& \multicolumn{1}{c|}{\rotatebox[origin=c]{90}{\makecell{Semantic Pointer \\ Architecture Unified Network}}}
& \multicolumn{1}{c|}{\rotatebox[origin=c]{90}{\makecell{Associative-Projective \\ Neural Networks}}}
& \multicolumn{1}{c|}{\rotatebox[origin=c]{90}{\makecell{Hierarchical \\ Temporal Memory}}}
& \multicolumn{1}{c|}{\rotatebox[origin=c]{90}{\makecell{Learning Intelligent \\ Distribution Agent}}}
& \multicolumn{1}{c|}{\rotatebox[origin=c]{90}{\makecell{Memories for \\ cognitive  architectures}}}
\\  \hline

     \cite{PlateCommon1997} & {\color{red} \xmark} & {\color{red} \xmark} & {\color{red} \xmark}  & {\color{red} \xmark} & {\color{red} \xmark} & {\color{red} \xmark}  & {\color{red} \xmark} & {\color{red} \xmark} & {\color{red} \xmark} & $\bm{\pm}$  & $\bm{\pm}$ & {\color{red} \xmark}  & {\color{red} \xmark} & {\color{red} \xmark} & {\color{red} \xmark} & {\color{red} \xmark} & {\color{red} \xmark}  & {\color{red} \xmark} \\  \hline 
     
     \cite{KanervaHyperdimensional2009} & {\color{red} \xmark} & {\color{red} \xmark} & {\color{red} \xmark} & $\bm{\pm}$ & {\color{red} \xmark} & {\color{red} \xmark}   & {\color{red} \xmark} & {\color{red} \xmark} & {\color{red} \xmark} & $\bm{\pm}$   & $\bm{\pm}$ & {\color{red} \xmark}  & {\color{red} \xmark} & {\color{red} \xmark} & {\color{red} \xmark} & {\color{red} \xmark} & {\color{red} \xmark}  & {\color{red} \xmark} \\  \hline 
     
     \cite{RahimiNanoscalable2017}& {\color{red} \xmark} & {\color{red} \xmark} & {\color{red} \xmark} & {\color{red} \xmark} & {\color{red} \xmark} & {\color{red} \xmark}   & $\bm{\pm}$ & {\color{red} \xmark} & {\color{red} \xmark} & {\color{red} \xmark}  & {\color{red} \xmark} & {\color{red} \xmark}  & {\color{red} \xmark} & {\color{red} \xmark} & {\color{red} \xmark} & {\color{red} \xmark} & {\color{red} \xmark}  & {\color{red} \xmark} \\  \hline 
     
     \cite{RahimiBiosignal2019} & {\color{red} \xmark} & {\color{red} \xmark} & {\color{red} \xmark}  & {\color{red} \xmark} & {\color{red} \xmark} & {\color{red} \xmark}   & $\bm{\pm}$ & {\color{red} \xmark} & {\color{red} \xmark} & {\color{red} \xmark} & {\color{red} \xmark} & {\color{red} \xmark}  & {\color{red} \xmark} & {\color{red} \xmark} & {\color{red} \xmark} & {\color{red} \xmark} & {\color{red} \xmark}  & {\color{red} \xmark}  \\  \hline     
     
     \cite{NeubertRobotics2019}& {\color{red} \xmark} & {\color{red} \xmark} & {\color{red} \xmark}  & {\color{red} \xmark} & {\color{red} \xmark} & $\bm{\pm}$  & $\bm{\pm}$ & $\bm{\pm}$ & {\color{red} \xmark} & {\color{red} \xmark}  & {\color{red} \xmark} & {\color{red} \xmark}  & $\bm{\pm}$ &{\color{red} \xmark} & {\color{red} \xmark} & {\color{red} \xmark} & {\color{red} \xmark}  & {\color{red} \xmark} \\  \hline        
     
     \cite{GeClassificationReview2020} & {\color{red} \xmark} & {\color{red} \xmark} & {\color{red} \xmark}  & {\color{red} \xmark} & {\color{red} \xmark} & {\color{red} \xmark}  & {\color{cadmiumgreen} \cmark} & {\color{red} \xmark} & {\color{red} \xmark} & {\color{red} \xmark}  & {\color{red} \xmark} & {\color{red} \xmark}  & {\color{red} \xmark} & {\color{red} \xmark} & {\color{red} \xmark} & {\color{red} \xmark} & {\color{red} \xmark}  & {\color{red} \xmark} \\  \hline

     \cite{SchlegelVSAComparison2020} & {\color{red} \xmark} & {\color{red} \xmark} & {\color{red} \xmark}   & {\color{red} \xmark} & {\color{red} \xmark} &  $\bm{\pm}$  & {\color{red} \xmark} & $\bm{\pm}$ & {\color{red} \xmark}& {\color{red} \xmark} & {\color{red} \xmark} & {\color{red} \xmark}  & $\bm{\pm}$ & {\color{red} \xmark} & {\color{red} \xmark} & {\color{red} \xmark} & {\color{red} \xmark}  & {\color{red} \xmark}  \\  \hline     
     
     \cite{KleykoComputingParadigm2021} & $\bm{\pm}$ & {\color{red} \xmark} & {\color{red} \xmark}  & {\color{red} \xmark} & {\color{red} \xmark} & {\color{red} \xmark}   & {\color{red} \xmark} & {\color{red} \xmark} & {\color{red} \xmark}& {\color{red} \xmark} & {\color{red} \xmark} & {\color{red} \xmark}  & {\color{red} \xmark} & {\color{red} \xmark} & {\color{red} \xmark} & {\color{red} \xmark} & {\color{red} \xmark}  & {\color{red} \xmark} \\  \hline   
     
     \cite{hassan2021hyper} & {\color{red} \xmark} & {\color{red} \xmark} & {\color{red} \xmark} & {\color{red} \xmark} & {\color{red} \xmark} & {\color{red} \xmark}   & {\color{cadmiumgreen} \cmark} & {\color{red} \xmark} & {\color{red} \xmark}& {\color{red} \xmark} &   {\color{red} \xmark} & {\color{red} \xmark}  & $\bm{\pm}$  & {\color{red} \xmark} & {\color{red} \xmark} & {\color{red} \xmark} & {\color{red} \xmark}  & {\color{red} \xmark}  \\  \hline   

     \hline    \makecell{This survey, Part II \\ Section \#} & \ref{sec:stor:automschemes} & \ref{sec:stor:trans:comm} & \ref{sec:stor:string} & \ref{sec:context:HVs} & \ref{sec:context:biomedical} & \ref{sec:context:images} & \ref{sec:app:class:features}
& \ref{sec:app:class:images}
& \ref{sec:app:class:structured}
& \ref{sec:database:holistic}
& \ref{sec:analogical:reasoning}
& \ref{sec:cog:modeling}
& \ref{sec:comp:vision}
& \ref{sec:cog:spaun}
& \ref{sec:cognitive:APNN}
& \ref{sec:cognitive:HTM}
& \ref{sec:cognitive:LIDA}
& \ref{sec:cognitive:memory}
     \\  \hline  
     
     \end{tabular}
    \end{center}
\end{table}

There were no previous attempts to make a comprehensive survey of HDC/VSA, but there are articles that overview particular topics of HDC/VSA. 
Probably the first attempt to overview and unify different HDC/VSA models should be attributed to Plate~\cite{PlateCommon1997}. 
The key idea for the unification was to consider the existing (at that time, four) HDC/VSA models as different schemes for implementing two key operations: binding and superposition (see Section~\ref{PartI-sec:vsa:operations} in~\cite{KleykoSurveyVSA2021Part1}). However, since that time numerous HDC/VSA models have come to prominence.
A more recent summary of the most frequently used models was provided in~\cite{RahimiNanoscalable2017}.
In~\cite{SchlegelVSAComparison2020}, the HDC/VSA models were compared in terms of their realizations of the binding operation. 
Both articles, however, missed some of the models.
These and other gaps have been filled in Part~I of this survey~\cite{KleykoSurveyVSA2021Part1}.

As for applications of HDC/VSA -- the topic covered in this article -- in Table~\ref{table:position:part2} we identified the following substantial application domains, which reflect the structure of Sections~\ref{sec:applications} and~\ref{sec:cog}: deterministic behavior, similarity estimation, classification, cognitive computing, and cognitive architectures.
The columns in Table~\ref{table:position:part2} list more fine-grained application clusters within these larger domains.

There is no previous article, which would account for all currently known applications, though there are recent works overviewing either a particular application area (as in~\cite{RahimiBiosignal2019}, where the focus was on biomedical signals), or certain application types (as in~\cite{GeClassificationReview2020, hassan2021hyper}, where solving classification tasks with HDC/VSA was the main theme).
The topic of machine learning is also omnipresent in this survey, and due to its ubiquity we dedicated Section~\ref{sec:app:class} to classification tasks. 
However, the scope of the survey is much broader as it touches on all currently known applications.  
Table~\ref{table:position:part2} contrasts the coverage of Part~II of this survey with the previous articles (ordered chronologically). We use $\bm{\pm}$ to indicate that an article partially addressed a particular topic, but either new results were reported since then or not all related work was covered.

In Part~I of this survey~\cite{KleykoSurveyVSA2021Part1}, we considered the motivation behind HDC/VSA and basic notions,  summarized currently known HDC/VSA models, and presented the transformation of various data types into HVs.
Part~II of this survey covers existing applications (Section~\ref{sec:applications})
and the use of HDC/VSA in cognitive modeling and architectures (Section~\ref{sec:cog}). 
The discussion and challenges, as well as conclusions, are presented in Sections~\ref{sec:disc} and  Section~\ref{sec:conc}, respectively.

\section{Application areas}
\label{sec:applications}

HDC/VSA have been applied across different fields for various  tasks.   
For this section, we aggregated the existing applications into several groups:
deterministic behavior (Section~\ref{sec:stor:trans}), similarity estimation (Section~~\ref{sec:apps:similarity}), and classification (Section~~\ref{sec:app:class}).

\subsection{Deterministic behavior with HDC/VSA}
\label{sec:stor:trans}

In this section, we consider several use-cases of HVs designed to produce some kind of deterministic behavior. Note that due to the capacity limitations of HVs (see Section~\ref{PartI-sec:capacity} in~\cite{KleykoSurveyVSA2021Part1}), achieving deterministic behavior depends on several design choices.
These include the dimensionality of HVs as well as, e.g., the number of atomic HVs and the kind of rules used for constructing compositional HVs, such as the number of arguments in the superposition operation. Note also that strictly speaking, not all application areas listed here are perfectly deterministic (in particular, communications in Section~\ref{sec:stor:trans:comm}), but the determinism is a desirable property in all the areas collected in this section.

\subsubsection{Automata, instructions, and schemas}
\label{sec:stor:automschemes}

\paragraph{Finite-state automata and grammars}
\label{sec:stor:automata}

A deterministic finite-state automaton is specified by defining a finite set of states, a finite set of allowed input symbols, a transition function (defines all transitions in the automaton), a start state, and a finite set of accepting states. The current state can change in response to an input. The joint current state and input symbol uniquely determine the next state of the automaton. 

\begin{figure}[t]
\centering
\includegraphics[width=0.35\columnwidth]{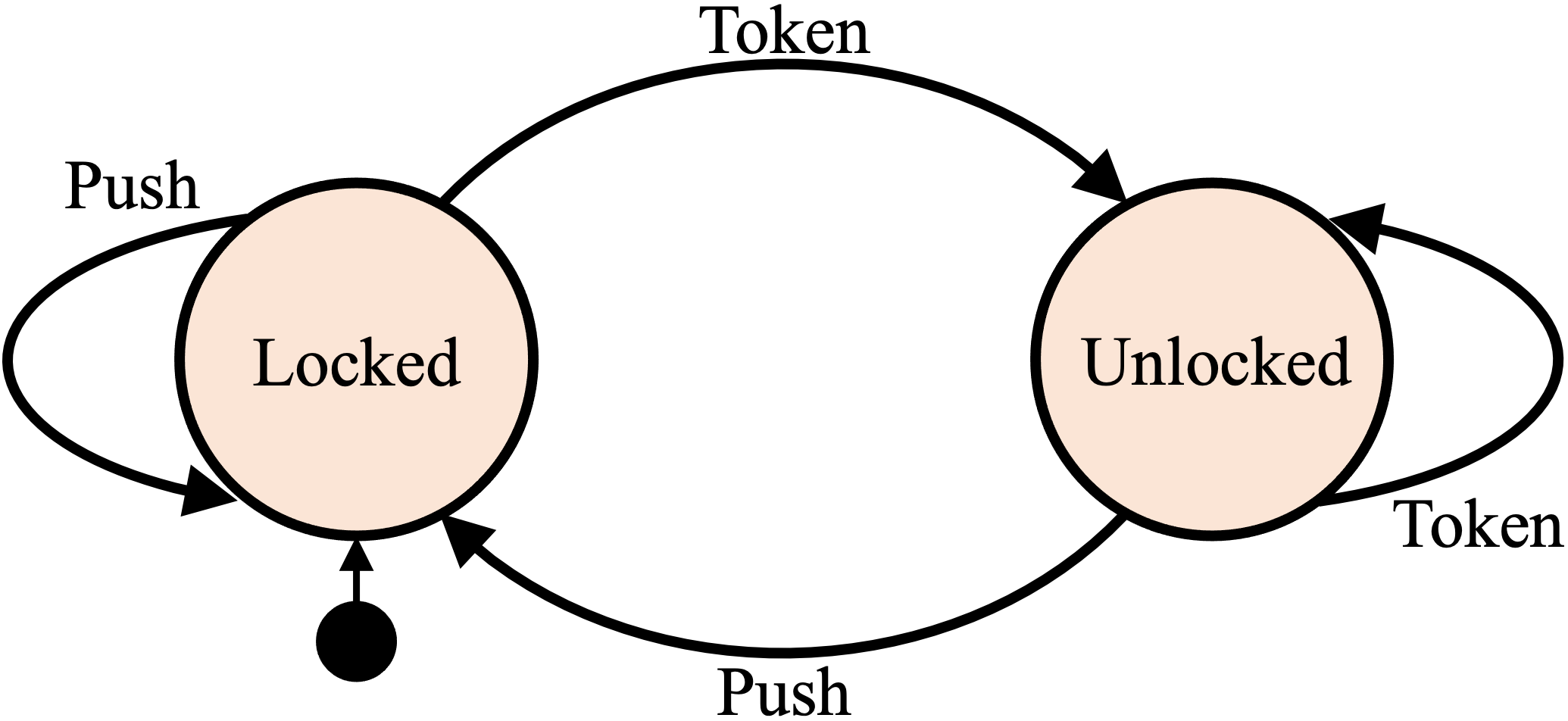}
\caption{
An example of a state diagram of a finite-state automaton modeling the control logic of a turnstile.
}
\label{fig:fsa}
\end{figure}

An intuitive example of an automaton controlling the logic of a turnstile is presented in Fig.~\ref{fig:fsa}. The set of input symbols is \{ ``Token'', ``Push''  \} and the set of states is \{ ``Unlocked'', ``Locked'' \}. The state diagram in Fig.~\ref{fig:fsa} can be used to derive the transition function.

HDC/VSA-based implementations of finite-state automata were proposed in~\cite{OsipovHD_FSA2017, YerxaUCBHD_FSA2018}. Random HVs are assigned to represent states ($\mathbf{u}$ for ``Unlocked''; $\mathbf{l}$ for ``Locked'') and input symbols ($\mathbf{t}$ for ``Token''; $\mathbf{p}$ for ``Push''). These HVs are used to form a compositional HV $\mathbf{a}$ for the transition function. The transformation is similar to the one used for the directed graphs in Section~\ref{PartI-sect:graphs:undirected} in~\cite{KleykoSurveyVSA2021Part1}. 
However, the HV representing the input symbol for the automaton is bound
to the edge HV that corresponds to the binding of the HVs for the current and the next state.
For instance, going from  ``Locked'' to ``Unlocked'' upon receiving ``Token'', is represented as: 
\noindent
\begin{equation}
\mathbf{t} \circ  \mathbf{l} \circ \rho(\mathbf{u}).
\end{equation}
\noindent
Given the HVs of all transitions, the transition function $\mathbf{a}$ is represented as their superposition: 
\noindent
\begin{equation}
\mathbf{a} =   
\mathbf{p} \circ \mathbf{l} \circ \rho(\mathbf{l}) + 
\mathbf{t} \circ \mathbf{l} \circ \rho(\mathbf{u}) + 
\mathbf{p} \circ \mathbf{u} \circ \rho(\mathbf{l}) + 
\mathbf{t} \circ \mathbf{u} \circ \rho(\mathbf{g}).
\end{equation}
\noindent

The next state is obtained by querying $\mathbf{a}$ with the binding of HVs of the current state and of the input symbol, followed by the inverse permutation of the resultant HV that returns the noisy version of the next state's HV\footnote{
It is worth recalling that this and other applications use a common design pattern, relying on the unbinding operation (see Section~\ref{PartI-sec:binding} in~\cite{KleykoSurveyVSA2021Part1}) that allows recovering one of the arguments. 
In the case of permutations, it is due to the fact that $\mathbf{a} = \rho^{-i}( \rho^{i}(\mathbf{a}))$, while in the case of multiplicative binding $\mathbf{a}= \mathbf{b} \oslash (\mathbf{a} \circ \mathbf{b})$, where for the implementations with self-inverse binding $\oslash = \circ$.   
}. 
For example, if the current state is $\mathbf{l}$ and $\mathbf{p}$ is received, then:
\noindent
\begin{equation}
\rho^{-1}(\mathbf{a} \circ \mathbf{p}  \circ \mathbf{l} )  =  \mathbf{l} +  \mathrm{noise}. 
\end{equation}
\noindent
This noisy HV is used as the query for the item memory to obtain the noiseless atomic HV $\mathbf{l}$.

Transformations of pushdown automata and context-free grammars into HVs have been presented in~\cite{GrabenGrammars2020}. 
A proposal for implementing Turing machines and cellular automata was given in~\cite{KleykoComputingParadigm2021}.

In~\cite{KnightGrammar2015}, the Holographic Reduced Representations (HRR) model was used to represent Fluid Construction Grammars, which is a formalism that allows designing construction grammars and using them for language parsing and production. Another work related to parsing is~\cite{StewartSentenceProcessing2014}, which presented an HDC/VSA implementation of a general-purpose left-corner parsing with simple grammars. An alternative approach to parsing with HVs using a constraint-based parser has been presented in~\cite{BlouwConstraintParsing2015}.

A related research direction was initiated by beim Graben and colleagues~\cite{beimGrabenGeometric12,GrabenGrammars2020,beimGrabenBook2012,WolffFockToolbox2018}. 
It concerns establishing a mathematically rigorous formalism of Tensor Product Representations using the concept
of a Fock space~\cite{Fock32}. 
The studies focus largely on  use-cases in computational linguistics, semantic processing, and quantum logic. 
The usage of the Fock space formalism for formulating minimalist grammars was presented in~\cite{beimGrabenGeometric12}.  Syntactic language processing as part of phenomenological modeling was reported in~\cite{beimGrabenBook2012}. 
An accessible and practical entry point to the area can be found in~\cite{WolffFockToolbox2018}.

\paragraph{Controllers, instructions, schemas}

In~\cite{LevyLearningBehavior2013}, using the Multiply-Add-Permute model (MAP), it was demonstrated how to manually construct a compositional HV that implements a simple behavior strategy of a robot. Sensor inputs as well as actions were represented as atomic HVs. Combinations of sensor inputs as well as combinations of actions were represented as a superposition of the corresponding atomic HVs. HVs of particular sensor input combinations were bound to the corresponding HV of action combinations. The bound HVs of all possible sensor-action rules were superimposed to produce the compositional HV of the robot controller. Unbinding this HV with the current HV of sensor input  combinations results in the noisy HV of the proper action combination. 
This idea was extended further in~\cite{NeubertRobotBehaviours2016} through a proposed algorithm to ``learn'' a compositional HV representing a robot's controller using the sensor-actuator values obtained from successful navigation runs.
This mode of robot operation is known as ``learning by demonstration''. It was realized as the superposition of the current ``controller'' HV with the HV corresponding to the binding of the sensor-actuator values - in case the current sensor HV was dissimilar to the ones already present in the ``controller'' HV. 
Another work studying robot navigation is~\cite{MenonNavigation2022}, which investigated a number of ways to form compositional HVs representing sensory data and explored the integration of the resultant HVs with neural network instead of the ``controller''.

In~\cite{ChooInstructionFollowing2013}, using HRR, instructions were represented as a sequence of rules, and rules as a sequence of their antecedent and consequent elements. Multiplicative bindings with position HVs were used to represent the sequence. Antecedents and consequents, in turn, were represented as HVs of their elements using binding and superposition operations. This approach was used as a part of instruction parsing in a cognitive architecture~\cite{StewartParsingSequentially2013}. 
In~\cite{LaihoSparse2015}, a proposal was sketched for a HDC/VSA-based processor, where both data and instructions were represented as HVs.

In~\cite{NeubertHDLearnPlan2018}, HVs for the representation of ``schemas'' in the form $<$\textit{context}, \textit{action}, \textit{result}$>$ were used. A more general approach for modeling the behavior of intelligent agents as ``functional acts''  was considered in~\cite{RachkovskijWorldModel2013} for  Sparse Binary Distributed Representations (SBDR; see also Section~\ref{sec:cognitive:APNN}). 
It is based on HVs representing triples $<$\textit{current situation}, \textit{action}, \textit{resulting situation}$>$ (which essentially correspond to ``schemas''), with the associated evaluations and costs. 
Finally, it is worth recalling that, in general, data structures to be represented by HVs do not have to be limited to ``schemas''.  
For example, a recent proposal in~\cite{GallantVSAJSON2022} suggested that HVs are well suited for forming representations of the JSON format that can include several levels of hierarchy.

\paragraph{Membership query and frequency estimation}

Section~\ref{PartI-sec:sets} in~\cite{KleykoSurveyVSA2021Part1} presented the transformation of sets and multisets into HVs. 
When implemented with the SBDR model, it becomes evident that Bloom Filters~\cite{Bloom1970space} are a special case of HDC/VSA~\cite{KleykoABF2020}, which have been used in a myriad of applications involving the membership query.
It is beyond the scope of this survey to overview them all, therefore, the interested readers are refereed to a survey in~\cite{TarkomaPracticeBF2012}).
When implementing the transformation of multisets via the Sparse Block Codes model, a similar connection can be made to count-min sketch~\cite{CormodeCountMin2005} that is used commonly for estimating frequency distributions in data streams (some applications are presented in~\cite{CormodeCountMin2005}).     
The use of the HDC/VSA principles for constructing hash tables has been recently considered in~\cite{HeddesHDHash2022}.

\subsubsection{Transmission of data structures}
\paragraph{Communications}
\label{sec:stor:trans:comm}

The main motivation for using HDC/VSA in the communication context is their robustness to noise due to the distributed nature of HVs. Let us consider three similar but varying applications of HDC/VSA.

In~\cite{JakimovskiCollective2012}, it was shown how to use Binary Spatter Codes (BSC) (Section~\ref{PartI-sec:vsa:bsc} in~\cite{KleykoSurveyVSA2021Part1}) in collective communication for sensing purposes. Multiple devices wirelessly sent their specific HVs representing some of their sensory data (the paper used temperature as a show-case). 
It was proposed to receive them in a manner that implements the superposition operation. This superposition HV was then analyzed by calculating the $\text{dist}_{\text{Man}}$ of the normalized superposition and the atomic HVs and comparing with a threshold. For instance, the case that a particular temperature was transmitted could be detected. Another version of analysis allowed checking how many devices had been exposed to a particular temperature. The proposed communication scheme does not require a control mechanism for getting multiple access to the medium. So, it can be useful in scenarios where there are multiple devices that have to report their states to some central node. 
Recently, it has been shown how such over-the-air superposition can be used for on-chip communications to scale up the architecture with multiple transmitters and receivers~\cite{WirelessGuirado2022}. 
This has been done by carefully engineering modulation constellations, and it paves the way for a large number of \emph{physically distributed} associative memories (as wireless-augmented receivers) to reliably perform the similarity search given a slightly different version of a query HV as their input.

In~\cite{KleykoMACOM2012}, BSC was used in the context of medium access control protocols for wireless sensor networks. A device was forming a compositional HV representing the device's sensory data (by the superposition of multiplicative bindings), which was then transmitted to the communication medium. It was assumed that the receiver knew the atomic HVs and, thus, could recover the information represented in the received compositional HV. The application scope of this approach is for scenarios where the communication medium is very harsh so that a high redundancy of HVs is useful for reliably transmitting the data.

In~\cite{KimHDM2018}, it was proposed to combine forward error correction and modulation using Fourier HRR (Section~\ref{PartI-sec:vsa:fhrr} in~\cite{KleykoSurveyVSA2021Part1}). The scheme represented individual pieces of data by using complex-valued HVs that were then combined into a compositional HV using permutation and superposition operations. Unnormalized complex values of the compositional HV were transmitted to the communication medium. The iterative decoding of the received compositional HV significantly increased the code rate. 
The application scope of this scheme is robust communication in a low signal-to-noise ratio regime. The scheme at a lower coding rate was compared to the low-density parity check and polar codes in terms of achieved bit error rates, while featuring lower decoding complexity. 
To improve its signal-to-noise ratio gain, a soft-feedback iterative decoding was proposed~\cite{HerscheHDMClassifier2021} to additionally take the estimation's confidence into account. That improved the signal-to-noise ratio gain by 0.2\,dB at a bit-error-rate of $10^{-4}$.
In further works, the scheme has been applied to collision-tolerant narrowband communications~\cite{HsuCollisionTolerant2019}, massive machine type communications~\cite{HsuNonOrthogonalModulation2020}, and near-channel classification~\cite{HerscheHDMClassifier2021}.

\paragraph{Distributed orchestration}

Another use-case of HDC/VSA in the context of transmission of data structures is distributed orchestration. The key idea presented in~\cite{SimpkinScalable2018, SimpkinHDWorkflow2019} was to use BSC to communicate a workflow in a decentralized manner between the devices involved in the application described by the workflow. Workflows were represented and communicated as compositional HVs constructed using the primitives for representing sequences (Section~\ref{PartI-sec:sequences} in~\cite{KleykoSurveyVSA2021Part1}) and directed acyclic graphs (Section~\ref{PartI-sect:graphs:labelled} in~\cite{KleykoSurveyVSA2021Part1}).
In~\cite{SimpkinHDWorkflow2020}, the approach was implemented in a particular workflow system -- Node-RED.
In~\cite{BarclayTrustable2022}, the approach was extended further to take into account the level of trust associated with various elements when selecting services.

\subsubsection{String processing}
\label{sec:stor:string}

In \cite{RachkovskijApplication1996}, it was proposed to obtain HVs of a word using permutations (cyclic shifts) of HVs of its letters to associate a letter with its position in the word. The conjunction was used to bind together all the obtained letter-in-position HVs. HVs of words formed in such a way were then used to obtain $n$-grams of sequences of words using the same procedure. The obtained HVs were used to estimate the frequencies of various word sequences in texts in order to create a model of human text reader interests. Note that here the conjunction result is somewhat similar to all input HVs. 

An interesting property of sequence representation using permutations of its element HVs (Section~\ref{PartI-sec:sequences} in~\cite{KleykoSurveyVSA2021Part1}) is that the HV of a shifted sequence can be obtained by the permutation of the sequence HV as a whole~\cite{KleykoSubstrings2014,MitrokhinSensorimotor2019,KleykoCommentariesSR2020}. This property was leveraged in~\cite{KleykoSubstrings2014} for searching the best alignment (shift) of two sequences, i.e., the alignment that provides the maximum number of coinciding symbols. This can be used, e.g., for identifying common substrings.
Such a representation, however, does not preserve the similarity of symbols' HVs in nearby positions, which would be useful for, e.g., spell checking. This can be addressed by, e.g., extending the permutation-based representations as in~\cite{KleykoPermuted2016}, where the resultant compositional HVs were evaluated on a permuted text, which was successfully reconstructed. 
An approach to transforming sequences into sparse HVs from~\cite{RachkovskijEquivariant2021}, which preserves the similarity of symbols at nearby positions and is shift-equivariant, was applied to the spellchecking task.

An algorithm for searching a query string in the base string was proposed in~\cite{PashchenkoSubstring2020} and modified in~\cite{KleykoComputingParadigm2021}. It is based on the idea of representing finite-state automata in an HV (see Section~\ref{sec:stor:automata}). The algorithm represented the base string as a non-deterministic finite-state automaton~\cite{RabinFinite1959}. The symbols of the base string corresponded to the transitions between the states of the automaton. The automaton in turn was represented as a compositional HV. The automaton was initialized as a superposition of all atomic HVs corresponding to the states. The query substring was presented to the automaton symbol by symbol. If after the presentation of the whole substring the automaton appeared in one of the valid states, this indicated the presence of the query substring in the base string. 

In \cite{KimEfficientDNA2020}, the MAP model was used for DNA string matching. The key idea was as in~\cite{JoshiNgrams2016}: the base DNA string was represented by one or several HVs containing a superposition of all $n$-grams of predefined size(s). HVs of $n$-grams were formed by multiplicative binding of appropriately permuted HVs of their symbols (Section~\ref{PartI-sec:sequences:ngrams} in~\cite{KleykoSurveyVSA2021Part1}). A query string was considered present in the base DNA string if the similarity between its HV and the compositional HV(s) of the base DNA string was higher than the predefined threshold. The threshold value determined the balance between true and false positives, similar to Bloom filters (see Section~\ref{PartI-sec:sets} in~\cite{KleykoSurveyVSA2021Part1}). The approach was evaluated on two databases of DNA strings: escherichia coli and human chromosome $14$. 
The main promise of the approach in \cite{KimEfficientDNA2020} is the possibility to accelerate string matching with application-specific integrated circuits due to the simplicity and parallelizability of the HDC/VSA operations.

\subsubsection{Factorization}
The resonator networks~\cite{FradyResonator2020, KentResonatorNetworks2020} were proposed as a way to solve factorization problems, which often emerge during the recovery procedure, within HDC/VSA (Section~\ref{PartI-sec:recovery} in~\cite{KleykoSurveyVSA2021Part1}).
It is expected that the resonator networks will be useful for solving various factorization problems, but this requires formulating the problem for the HDC/VSA domain. 
An initial attempt to decompose synthetic scenes was demonstrated in~\cite{FradyDisentangling2018}. A more recent formulation for the integer factorization was presented in~\cite{KleykoPrimes2022}.
The transformation of numbers into HVs was based on the fractional power encoding~\cite{FradyFunctions2021, FradyFunctionsNICE2022} (Section~\ref{PartI-sec:scalars:vectors:compos} in~\cite{KleykoSurveyVSA2021Part1}) combined with log transformation, and the resonator network was used to solve the integer factorization problem.
The approach was evaluated on the factorization of semiprimes.

\subsection{Similarity estimation with HVs}
\label{sec:apps:similarity}

Transformations of original data into HVs allow constructing HVs in various application areas in a manner that preserves similarity relevant for a particular application. 
This provides a tool to use HDC/VSA for ``similarity-based reasoning''
that includes, e.g., similarity search and classification in its simplest form as well as the much more advanced analogical reasoning considered in Section~\ref{sec:analogical:reasoning}.
Due to the abundance of studies on classification tasks we devote another separate Section~\ref{sec:app:class} to it. 
In this section, we primarily focus on ``context'' HVs, since for a long time these were the most influential application of HDC/VSA. We also present some existing efforts in similarity search.

\subsubsection{Word embeddings: context HVs for words and texts}
\label{sec:context:HVs}

The key idea for constructing  context vectors is usually referred to as the
``distributional semantics hypothesis''~\cite{HarrisLanguage1968}, suggesting that linguistic items with similar distributions have similar meanings. The distributions are calculated as frequencies of item occurrence in particular contexts by using a document corpus. For items-words the contexts could be, e.g., documents, paragraphs, sentences, or sequences of words close to a focus word. For example, the generalized vector space model~\cite{WongModeling1987} used documents as contexts for words in information retrieval. In computational linguistics and machine learning, vectors are also known as embeddings. 

In principle, context vectors can be obtained in any domain where objects and contexts could be defined. Below, we will only focus on the context vector methods which are commonly attributed to HDC/VSA. They usually transform the frequency distributions into HVs of particular formats, which we will call context HVs. 

Historically, the first proposal to form context HVs for words was that of Gallant, e.g.,~\cite{GallantPractical1991, GalantContext1993, CaidLearned1995, GalantContext2000}.
These studies, however, did not become known widely, but see~\cite{HechtContext1994}.
In fact, two most influential HDC/VSA-based methods for context HVs are Random Indexing (RI)~\cite{KanervaRI2000, SahlgrenRIIntroduction2005} and 
Bound Encoding of the Aggregate Language Environment (BEAGLE)~\cite{JonesMeaning2007}.

\begin{table}[t]
\tiny
\renewcommand{\arraystretch}{1.0}
\caption{Experiments with context vectors.}
\centering
\label{tab:app:cont:vec}
\begin{tabular}{|c|c|c|c|c|}
\hline
Ref. & Task & Dataset & Method  & Baseline(s) \\ 
\hline

\cite{KanervaRI2000} & \makecell{Synonymy test} &  \makecell{TOEFL}     & \makecell{RI with word-document matrix} 
 &  LSA \\ \hline

\cite{SahlgrenVector2001} & \makecell{Synonymy test} &  \makecell{TOEFL}     & \makecell{RI with word-word matrix} 
 &  LSA \\ \hline

\cite{MisunoVector2005} & \makecell{Synonymy test } &  \makecell{TOEFL; ESL }     & \makecell{RI with word-word matrix} 
 &  LSA, RI \\ \hline

\cite{MisunoVector2005} & \makecell{Semantic similarity of word pairs } &  \makecell{ \cite{RubensteinContextual1965}   }     & \makecell{RI with word-word matrix} 
 &  LSA, RI \\ \hline
 
 \cite{MisunoVector2005} & \makecell{ Word choice in  Russian to English translation} &  \makecell{Own data}     & \makecell{RI with word-word matrix} 
 &  LSA, RI \\ \hline

\cite{MisunoSearching2005} & \makecell{Semantic text  search} & \makecell{MEDLARS; Cranfield; Time Magazine}     & \makecell{RI with word-document matrix} 
 &   \makecell{(Generalized) Vector   Space Model} \\ \hline

\cite{SahlgrenOrder2008} & \makecell{Synonymy test} &  \makecell{TOEFL}    & \makecell{RI with permutations} 
 &  BEAGLE \\ \hline

\cite{CohenCancer2012} & \makecell{Retrieval of cancer therapies} &  \makecell{A set of predications\\ extracted from MEDLINE}    & \makecell{FHRR-based PSI}  &  BSC-based PSI \\ \hline

\cite{CohenPredictingScreening2014}  &
\makecell{Identification of agents \\ active against cancer cells}   &  
\makecell{SemMedDB}     & 
PSI &
Reflective RI  \\ \hline 
 
\cite{RecchiaSemantic2015} & \makecell{Synonymy test} &  \makecell{TOEFL; ESL}    & \makecell{BEAGLE;  RI with permutations} 
 &  BEAGLE \\ \hline
 
\cite{RecchiaSemantic2015} & \makecell{Semantic similarity of word pairs} &  \makecell{From \cite{RubensteinContextual1965} \& 
\cite{MillerContextual1991} \&  
\cite{ResnikSemantic1995} \&  
\cite{FinkelsteinSemantic2002} }    & \makecell{BEAGLE;  RI with permutations} 
 &  BEAGLE \\ \hline

\cite{JamiesonITS2018} & \makecell{Taxonomic organization} &  \makecell{From~\cite{JonesMeaning2007}  \&  \cite{LundProducing1996}}    & ITS 
 &  \makecell{BEAGLE;  LSA} \\ \hline

 \cite{JamiesonITS2018} & \makecell{Meaning disambiguation \ in context} &  \makecell{From~\cite{SchvaneveldtLexical1976}}    & ITS 
 &  \makecell{BEAGLE;  LSA} \\ \hline

\cite{JohnsExperiential2019} & \makecell{Synonymy test} &  \makecell{TOEFL}    & \makecell{BEAGLE with  experiential optimization} 
 &  \makecell{BEAGLE } \\ \hline

\cite{CohenInteractions2019}  &
\makecell{Prediction of side-effects \\ for drug combinations}  &  
\makecell{From \cite{ZitnikModeling2018}}     & 
ESP &
graph convolutional ANN  \\ \hline

\cite{JohnsExperiential2019} & \makecell{Semantic similarity  of word pairs} &  \makecell{From \cite{RubensteinContextual1965} \&  
\cite{MillerContextual1991} \&  
\cite{FinkelsteinSemantic2002} }    & \makecell{BEAGLE with  experiential optimization} 
 &  BEAGLE \\ \hline

\cite{aujla2019semantic}  &
\makecell{Academic search engine \\ for cognitive psychology}  &  
\makecell{Own data}     & 
BEAGLE &
RI with permutations  \\ \hline

\cite{johns2019influence} &
\makecell{Influence of corpus effects\\ on lexical behavior}  &  
\makecell{English Lexicon Project;\\British Lexicon Project; etc.}    & 
BEAGLE &
N/A  \\ \hline

\cite{SchubertLanguage2020}  &
\makecell{Contextual similarity among \\ alphanumeric characters}   &  
\makecell{Own data}    & 
RI with characters &
Word2vec~\cite{MikolovWord2vec2013} \& EARP~\cite{CohenEARP2018}  \\ \hline  

\cite{TalerFluency2020} & \makecell{Changes in verbal fluency} &  \makecell{Canadian Longitudinal  Study of Aging}    & BEAGLE 
 & N/A\\ \hline

\end{tabular}
\end{table}

\paragraph{Random Indexing}
\label{sec:random:indexing}

The RI method~\cite{KanervaRI2000, SahlgrenRIIntroduction2005}
was originally proposed in~\cite{KanervaRI2000} as a simple alternative to Latent Semantic Analysis (LSA)~\cite{LandauerSolution1997}. Instead of the expensive Singular Value Decomposition (SVD) used in LSA for the dimensionality reduction of a word-document matrix, RI uses the multiplication by a random matrix, thereby performing random projection (Section~\ref{PartI-sect:rand:proj} in~\cite{KleykoSurveyVSA2021Part1}). 
The Random Projection (RP) matrix was ternary ($\{-1, 0, 1\}$) and sparse.
Each row of the RP matrix is seen as a 
``random index'' (hence RI) assigned to each context (in that case, the document). In implementation, the frequency matrix was not formed explicitly, and the resultant context HVs were formed by scanning the document corpus and adding the document's random index vector to the context HV of each word in the document. 
The sign function can be used to obtain binary context HVs. 
The similarity between unnormalized context HVs was measured by $\text{sim}_{\text{cos}} $. 
The synonymy part of TOEFL was used as the benchmark to demonstrate that a performance comparable to LSA could be achieved, but at lower computational costs due to the lack of SVD.
In~\cite{PapadimitriouLatent2000}, it was proposed to speed up LSA by using RP before SVD as a preprocessing step.
In~\cite{KarlgrenWords2001}, similar to~\cite{GalantContext1993, LundProducing1996}, RI was modified to use a narrow context window consisting of only few adjacent words on each side of the focus word.

In~\cite{SahlgrenOrder2008}, permutations were used to represent the order information within the context window.
Further extensions of RI included generalization to multidimensional arrays (N-way RI)~\cite{SandinMultidimensionalRI2017} and inclusion of extra-linguistic features~\cite{KarlgrenUtterances2019}.
The RI was also extended to the case of training corpora that include information about relations between words. This model is called Predication-based Semantic Indexing (PSI)~\cite{WiddowsDiscoveryPatterns2012, CohenAnalogy2014, CohenSemantic2017}. PSI has been mainly used in biomedical informatics for literature-based discovery, such as identification of links between pharmaceutical substances and diseases they treat (see \cite{WiddowsDiscoveryPatterns2012} for more details). 
Later, PSI was extended to the Embedding of Semantic Predications (ESP) model that incorporates some aspects of ``neural'' word embeddings from~\cite{MikolovWord2vec2013} and similarity preserving HVs (Section~\ref{PartI-sec:scalars:vectors} in~\cite{KleykoSurveyVSA2021Part1}) for representing time periods~\cite{WiddowsGradedVectors2015}.

It is worth mentioning that the optimization-based method~\cite{SutorLifeLearning2018} for obtaining similarity preserving HVs from co-occurrence statistics can be contrasted to RI. RI also uses co-occurrence statistics, but  implicitly (i.e., without constructing the co-occurrence matrix). The difference, however, is that the RI is optimization-free and it forms HVs through a single pass via the training data. Thus, it can be executed online in an incremental manner, while the optimization required by~\cite{SutorLifeLearning2018} calls for iterative processing, which might be more suitable for offline operations.

 \paragraph{Bound Encoding of the Aggregate Language Environment}
\label{sec:BEAGLE}

There is an alternative method for constructing context HVs with HDC/VSA, known as BEAGLE~\cite{JonesMeaning2007}.
It was proposed independently of RI and used HRR to form the HVs of $n$-grams of words within a sentence, providing a representation of the order and context HVs of words. Words were initially represented with random atomic HVs as in HRR. 
The context HV of a word was obtained from two ``parts''. 
The first part included 
summing  the atomic HVs of the words in the sentence other than the focus word, for all corpus sentences.  
The second part was contributed by the word order HVs, which were formed as the superposition of the word $n$-gram HVs (with $n$ between $2$ and $7$). 
The $n$-gram HVs were formed with the circular convolution-based binding of the special atomic HV in place of the focus word and the atomic HVs of the other word(s) in the $n$-gram.
The word order in an $n$-gram was represented recursively, first by binding the HVs of the left and the right word permuted differently, and then by binding the resultant HV with the next right word, again using permutations for the ``left'' and ``right'' relative positions. 
The total context HV was the superposition of the HVs for the two parts. 
The similarity measure was $\text{sim}_{\text{cos}} $.

Later,~\cite{RecchiaSemantic2010} presented a modified version of BEAGLE using random permutations. The authors found that their model was both more scalable to large corpora and gave better fits to semantic similarity than the circular convolution-based representation. A comprehensive treatment of methods for constructing context HVs of phrases and sentences as opposed to individual words was presented in~\cite{MitchellComposition2010}.
In a similar spirit, in~\cite{LuoDecomposed2018} the HRR model was used to construct compositional HVs that were able to discover language regularities resembling syntax and semantics.
The comparison of different sentence embeddings (including BEAGLE) in terms of their ability to represent syntactic constructions was provided in~\cite{KellySentence2020}.
It has been demonstrated that context HVs formed by BEAGLE can account for a variety of semantic category effects such as typicality, priming, and acquisition of semantic and lexical categories. The effect of using different lexical materials to form context HVs with BEAGLE was demonstrated in \cite{JohnsExperiential2019}, while the use of negative information for context HVs was assessed in~\cite{JohnsNegative2019}. 
In~\cite{KellyDegrees2017}, BEAGLE was extended to a Hierarchical Holographic Model by augmenting it with additional levels corresponding to higher-order associations, e.g., part-of-speech and syntactic relations.
The Hierarchical Holographic Model was used further in~\cite{KellyIndirect2020} to investigate what grammatical information is available in context HVs produced by the model.

Due to high similarity between BEAGLE and RI, both methods were compared against each other in~\cite{RecchiaSemantic2010, RecchiaSemantic2015}. It was shown that both methods demonstrate similar results on a set of semantic tasks using a Wikipedia corpus for training. The main difference was that RI is much faster as it does not use the circular convolution operation.
For placing the methods in the general context of word embeddings, please refer  to~\cite{WiddowsSemanticComposition2021}.

The work \cite{MercadoSemantic2020} aimed to address the representation of similarity, rather than relatedness represented in context HVs by BEAGLE and RI. To do so, for each word the authors represented its most relevant semantic features taken from a knowledge base ConceptNet. The context HV of a word was formed using BSC as a superposition of its semantic feature HVs formed by role-filler bindings. The results from measuring  the  semantic similarity  between  pairs  of  concepts were presented using the SimLex-999 dataset.

Table~\ref{tab:app:cont:vec} provides a  non-comprehensive summary of the studies that applied the BEAGLE and RI methods to various linguistic tasks. 
The interested readers are referred to~\cite{WiddowsDistributionalSemantics2009}, which is a survey describing the applications of RI, PSI, and related methods, including the biomedical domain. The range of applications described in~~\cite{WiddowsDistributionalSemantics2009}  covers word-sense disambiguation, bilingual information extraction, visualization of relations between terms, and document retrieval.
It is  worth noting that there is a software package called ``Semantic vectors''~\cite{WiddowsSemanticInitial2008}, \cite{WiddowsSemanticVectors2008}, \cite{WiddowsSemVec2010}, that implements many of the methods mentioned above and provides the main building blocks for designing further modifications of the methods.

\subsubsection{Similarity estimation of biomedical signals}
\label{sec:context:biomedical}

In~\cite{KleykoBreathing2018, KleykoDeepBreathing2019}, BSC was applied to biomedical signals: heart rate and respiration. 
The need for comparing these signals emerged in the scope of a deep breathing test for assessing autonomic function. 
HDC/VSA was used to analyze  cardiorespiratory  synchronization by comparing the similarity between heart rate and respiration using feature-based analysis.   
Feature vectors were extracted from the signals and transformed into HVs by using role-filler bindings (Section~\ref{PartI-sec:key:value} in~\cite{KleykoSurveyVSA2021Part1}) and representations of scalars (Section~\ref{PartI-sec:scalars:vectors} in~\cite{KleykoSurveyVSA2021Part1}).
These HVs were in turn  classified into different degrees of cardiorespiratory synchronization/desynchronization. 
The signals were obtained from the healthy adult controls, patients with cardiac autonomic neuropathy, and patients with myocardial infarction. It was shown that, as expected, the similarity between different HVs  were lower for patients with the cardiac autonomic neuropathy and myocardial infarction patients, than for the healthy controls. 

Another application of BSC was the identification of the ictogenic (i.e., seizure generating) brain regions from intracranial electroencephalography (iEEG) signals~\cite{BurrelloLBPSeizure2020}. 
The algorithm first transformed iEEG time series from each electrode into a sequence of symbolic local binary pattern codes, from which a binary HV was obtained for each brain state (e.g., ictal or interictal). It then identified the ictogenic brain regions by measuring the relative distances between the learned HVs from different groups of electrodes. Such the identification was done by one-way ANOVA tests at two levels of spatial resolution, the cerebral hemispheres and lobes.

\subsubsection{Similarity estimation of images}
\label{sec:context:images}

In~\cite{NeubertAggregation2021}, the HDC/VSA 2D image representations (Section~\ref{PartI-sec:2Dimages} in~\cite{KleykoSurveyVSA2021Part1}) were applied for an aggregation of local descriptors extracted from images. Local image descriptors were real-valued vectors whose dimensionality was controlled by RP (Section~\ref{PartI-sect:rand:proj} in~\cite{KleykoSurveyVSA2021Part1}). To represent a position inside an interval, the authors concatenated parts of two basis HVs and used several intervals, as in~\cite{RachkovskijScalars2005, RachkovskijVectors2005}, but using MAP. Position HVs for \textit{x} and \textit{y} were bound to represent (\textit{x},\textit{y}), see Section~\ref{PartI-sec:2Dimages} in~\cite{KleykoSurveyVSA2021Part1}.
Subsequently, projected local image descriptors were bound with their position HVs, using component-wise multiplication, and the bound HVs were superimposed to represent the whole image. The image HVs obtained with different algorithms for extracting the descriptors could also be aggregated using the superposition operation. 
When compared to the standard aggregation methods in (mobile robotics) place recognition experiments, HVs of the aggregated descriptors exhibited an average performance better than alternative methods (except the exhaustive pair-wise comparison).
A very similar concept was demonstrated in~\cite{MitrokhinCNN2020} using an image classification task, see also Table~\ref{tab:class:images}. One of the proposed ways of forming image HV used the superposition of three binary HVs obtained from three different hashing neural networks.
The HVs representing the aggregated descriptors provided a higher classification accuracy. 
Finally, similarity-preserving shift-equivariant representation of images in HVs using permutations was proposed in~\cite{rachkovskij2022representation}.


\begin{table}[t]
\tiny
\renewcommand{\arraystretch}{1.0}
\caption{HDC/VSA studies classifying languages}
\centering
\label{tab:class:language}
\begin{tabular}{|c|c|c|c|c|c|c|}
\hline
Ref. & Task & Dataset & HV format & 
\makecell{Primitives used in  data transformation}
  & Classifier & Baseline(s) \\ 
\hline

\cite{JoshiNgrams2016} & Language identification  &  \makecell{Wortschatz Corpora \&  Europarl Corpus}    & bipolar & \makecell{binding; permutation; superposition}
 & \makecell{centroids} & \makecell{vector  centroids}\\ \hline

\cite{RahimiLPHD} & Language identification  &  \makecell{Wortschatz Corpora \&  Europarl Corpus}    & \makecell{dense  binary} & \makecell{binding; permutation; superposition}
 & \makecell{binarized  centroids} & \makecell{localist  centroids}\\ \hline

\cite{ImaniSparseLanguage2017} & Language identification  &  \makecell{Wortschatz Corpora \&  Europarl Corpus}    & \makecell{sparse  binary} & \makecell{binding; permutation; superposition}
 & \makecell{binarized  centroids} & \makecell{approach from  \cite{RahimiLPHD}}\\ \hline

\cite{KleykoBoostingSOM2019} & Language identification  &  \makecell{Wortschatz Corpora \&  Europarl Corpus}    &  bipolar & \makecell{binding; permutation; superposition}
 & \makecell{self-organizing  map} & approach from \cite{JoshiNgrams2016} \\ \hline

\cite{SangaliAssociativeRecover2020} & Language identification  &  \makecell{Wortschatz Corpora \&  Europarl Corpus}    & \makecell{dense  binary} & \makecell{binding; permutation; superposition}
 & \makecell{evolvable binarized centroids} & \makecell{fastText}\\ \hline

\end{tabular}
\end{table}

\begin{table}[t]
\tiny
\renewcommand{\arraystretch}{1.0}
\caption{HDC/VSA studies classifying texts}
\centering
\label{tab:class:texts}
\begin{tabular}{|c|c|c|c|c|c|c|}
\hline
Ref. & Task & Dataset & HV format & \makecell{Primitives used in  data transformation}  & Classifier & Baseline(s) \\ 
\hline

\cite{RachkovskijClassifiers2007} & \makecell{Text    classification}  &  \makecell{Reuters-21578 }    & \makecell{sparse  binary}  & \makecell{ thresholded RP   }
 & SVM & \makecell{SVM with
 frequency vectors} \\ \hline

\cite{fishbein2008integrating} & \makecell{News  classification}  &  \makecell{20 Newsgroups}    &  real-valued & \makecell{binding; superposition}
 & \makecell{SVM with context and part of speech HVs} & \makecell{SVM with  context HVs}\\ \hline

\cite{NajafabadiText2016} & \makecell{News  classification}  &  Reuters newswire    &  \makecell{dense  binary}   & \makecell{binding; permutation; superposition}
 & \makecell{centroids} & \makecell{Bayes,
$k$NN,  and SVM without HVs}\\ \hline

\cite{ShridharEnd2End2020} & Intent classification  &  \makecell{Chatbot;  Ask Ubuntu;  Web Applications}    &  \makecell{dense  binary} & \makecell{binding; permutation; superposition}
 & \makecell{binarized ANN} & \makecell{classifiers without HVs}\\ \hline

\cite{AlonsoHyperEmbed2020} & Intent classification  &  \makecell{Chatbot;  Ask Ubuntu;  Web Applications}     &  bipolar & \makecell{binding; permutation; superposition}
 & \makecell{machine learning  algorithms} & \makecell{machine learning  algorithms without HVs}\\ \hline

\cite{JiaSpamHD2021} & 
Text spam detection  &
\makecell{Hotel reviews; SMS text; YouTube comments} &  
bipolar & 
\makecell{binding; permutation; superposition} &
\makecell{refined centroids} & 
\makecell{$k$NN; SVM; ANN; Random Forest}\\ \hline

\end{tabular}
\end{table}

\begin{table}[t!]
\caption{HDC/VSA studies classifying acoustic signals}
\label{tab:class:acoustic}
\centering
\tiny
\begin{tabular}{|c|c|c|c|c|c|c|}
\hline
Ref. & Task & Dataset & HV format & \makecell{Primitives used in  data transformation}  & Classifier & Baseline(s) \\ 
\hline

\cite{RachkovskijAudio1990} & \makecell{Vowels    recognition}  &  \makecell{ Own data}    & \makecell{sparse   binary}  & {\makecell{binding;  superposition } }
 & \makecell{stochastic perceptron;  centroids in associative memory}  & \makecell{N/A} \\ \hline

\cite{RachkovskijClassifiers2007} & \makecell{Distinguish nasal    and oral sounds}  &  \makecell{Phoneme  dataset}    & \makecell{sparse   binary}  & \makecell{RSC or Prager\\ }
 & SVM & \makecell{ANN; $k$NN; IRVQ} \\ \hline

\cite{RasanenMultivariate2015} & \makecell{Recognition of  spoken words}  &  CAREGIVER Y2 UK    &  \makecell{real-valued}   & \makecell{RP; binding; permutation;  superposition}
 & \makecell{centroids} &  \makecell{Gaussian mixture-based model} \\ \hline

\cite{ImaniVoiceHD2017} & \makecell{Recognition of  spoken letters}  &  Isolet    &  \makecell{dense  binary}   & \makecell{ binding;  superposition}
 & \makecell{binarized centroids;  binarized centroids \& ANN} &  3-layer ANN \\ \hline

\cite{ImaniHierarchical2018} & \makecell{Recognition of  spoken letters}  &  Isolet    &  \makecell{dense  binary}   & \makecell{ binding; permutation;  superposition}
 & \makecell{refined centroids} &   \makecell{approach from  \cite{ImaniVoiceHD2017}} \\ \hline

\cite{WongUnconventional2018} & \makecell{Recognition of   music genres}  &  Own data   &  N/A   & \makecell{ superposition}
 & \makecell{centroids} &  N/A \\ \hline

\cite{ImaniAdaptHD2019} & \makecell{Recognition of  spoken letters}  &  Isolet    &  \makecell{dense  binary}   & \makecell{ binding;  superposition}
 & \makecell{binarized refined centroids} &   \makecell{binarized centroids} \\ \hline

\cite{ImaniMemoryCentric2020} & \makecell{Recognition of  spoken letters}  &  Isolet      & \makecell{dense binary} & binding; superposition & \makecell{ multiple binarized  refined centroids  }  & $k$NN \\ \hline

\cite{Fragoso20} &  \makecell{Speaker recognition}  &  \makecell{Own data}      & \makecell{sparse  binary}  &  \makecell{LIRA}
 & \makecell{large margin  perceptron} & \makecell{ N/A} \\ \hline

\cite{HernandezOnlineHD2021} & \makecell{Recognition of  spoken letters}  &  Isolet     & bipolar & binding; superposition & \makecell{ conditioned  centroids }  & ANN; SVM; AdaBoost \\ \hline

\cite{ZouManiHD2021} & \makecell{Recognition of  spoken letters}  &  Isolet     & N/A & \makecell{trainable projection matrix} & \makecell{ refined  centroids }  & ANN; SVM; AdaBoost \\ \hline

\cite{hsiao2021hyperdimensional} & \makecell{Recognition of  spoken letters}  &  Isolet     & bipolar &  \makecell{trainable projection  matrix} & \makecell{ binarized refined  centroids }  & \makecell{binarized  centroids} \\ \hline

\cite{kazemi2021mimhd} &  \makecell{Recognition of  spoken letters}  &  Isolet      & bipolar & binding; superposition & \makecell{ quantized   refined centroids }  & \makecell{ non-quantized refined centroids } \\ \hline

\cite{ChangMulTaHDC2021} & \makecell{Recognition of  spoken letters}  &  Isolet     & bipolar &  \makecell{binding; superposition} & \makecell{ binarized  centroids }  & \makecell{approach from \cite{ImaniSparseHD2019}} \\ \hline

\cite{ZhangRobustness2021} &  \makecell{Recognition of  spoken letters}  &  Isolet      & bipolar & binding; superposition & \makecell{ quantized refined centroids }  & \makecell{N/A} \\ \hline

\cite{PoduvalStocHD2021} & \makecell{Recognition of  spoken letters}  &  Isolet     & bipolar & permutation; superposition & \makecell{ binarized refined  centroids }  & ANN; SVM; AdaBoost \\ \hline

\cite{YuUnderstandingHDC2022}  &
\makecell{Recognition of  spoken letters}   &  
Isolet     & 
integer-valued &
binding; superposition  & 
\makecell{ discretized stochastic gradient descent}  & 
\makecell{approach from \cite{ImaniBinary2019}}  \\ \hline

\cite{NazemiICCAD2020} & 
\makecell{Recognition of spoken letters}  &  Isolet   & 
integer-valued & 
compact code by ANN at low dimension & \makecell{centroids}  & 
ANN; other HDC/VSA solutions \\ \hline

\cite{MaMultimodal2022} & 
\makecell{Multimodal \\ sentiment analysis}  &
\makecell{CMU MOSI; CMU MOSEI} &  
real-valued & 
\makecell{binding; weighted superposition} &
\makecell{multimodal transformer} & 
\makecell{LSTM; multimodal transformer}\\ \hline

\end{tabular}
\end{table}

\subsection{Classification}
\label{sec:app:class}

Applying HDC/VSA to classification tasks is currently one of the most common application areas of HDC/VSA. This is due to the fact that similarity-based and other vector-based classifiers are widespread and machine learning research is on the rise in general. The recent survey~\cite{GeClassificationReview2020} of classification with HDC/VSA was primarily devoted to the transformation of input data into HVs. Instead, here we focus first on the types of input data (in the 2nd level headings), and then on the domains where HDC/VSA have been applied (in the 3rd level headings). Moreover, we cover some of the studies not presented in~\cite{GeClassificationReview2020}.
The studies are summarized in the form of tables, where each table specifies a reference, type of task, dataset used, format of HVs, operations to form HVs from data\footnote{
For the sake of generality, it was decided to avoid going to in-depth details of data transformations, so tables only specify the HDC/VSA operations used to construct HVs from data. 
}, 
the type of classifier, and baselines for comparison.
For the sake of consistency, in this section we use a table even if there is only a single work in a particular domain.

\subsubsection{ Classification based on feature vectors}
\label{sec:app:class:features}

\paragraph{Language identification with the vector of \textit{n}-gram statistics of letters}

In~\cite{JoshiNgrams2016}, it was shown how to form a compositional HV corresponding to $n$-gram statistics (see Section~\ref{PartI-sec:sequences:ngrams} in~\cite{KleykoSurveyVSA2021Part1}). 
The work also introduced a task of identifying a language amongst $21$ European languages. 
Since then, the task was used in several studies summarized in Table~\ref{tab:class:language}.

\paragraph{Classification of texts }

Table~\ref{tab:class:texts} summarizes the efforts of using HDC/VSA for text classification. 
The works in this domain dealt with different tasks such as text categorization, news identification, and intent classification. 
Most of the works~\cite{RachkovskijClassifiers2007, ShridharEnd2End2020, AlonsoHyperEmbed2020} used HVs as a way of representing data for conventional machine learning classification algorithms.

\paragraph{Classification of feature vectors extracted from acoustic signals}

Classification of various acoustic signals using HVs is provided in Table~\ref{tab:class:acoustic}. Tasks were mainly related to speech recognition, e.g, recognition of spoken letters or words.

\paragraph{Fault classification}

Studies on applying HDC/VSA to fault classification are limited. 
We are only aware of two such use-cases (summarized in Table~\ref{tab:class:fault}) applied to the problems of anomaly detection in a power plant and ball bearings.
An earlier work on micro machine-tool acoustic diagnostics was presented in~\cite{KussulDiagnostics1998}. 

\begin{table}[t]
\tiny
\renewcommand{\arraystretch}{1.0}
\caption{HDC/VSA studies classifying faults}
\centering
\label{tab:class:fault}
\begin{tabular}{|c|c|c|c|c|c|c|}
\hline
Ref. & Task & Dataset & HV format & \makecell{Primitives used in  data transformation}  & Classifier & Baseline(s) \\ 
\hline

\makecell{\cite{KussulDiagnostics1998}} & \makecell{Acoustic diagnostics
of micro machine-tools}  &  \makecell{Own data}     &  \makecell{sparse  binary}   & \makecell{RSC}
 & \makecell{large margin  perceptron} & N/A\\ \hline

\makecell{ \cite{KleykoFaultDetection2015}; \cite{KleykoIndustrial2018}  } & \makecell{Fault isolation}  &  \makecell{From \cite{PapakonstantinouSimulation2014}}     &  \makecell{dense  binary}   & \makecell{binding;  superposition}
 & \makecell{average $\text{dist}_{\text{Ham}}$ to the  training data HVs} & $k$NN\\ \hline

\makecell{\cite{EggimannConfigurableHD2021}} & \makecell{Ball bearing   anomaly detection}  & \makecell{IMS Bearing  Dataset}   & \makecell{dense  binary} & \makecell{binding; permutation;  superposition}  
 & \makecell{binarized  centroids} & N/A \\ \hline

\makecell{\cite{AmrouchWaferDefClass2021}} & 
\makecell{Detection of wafer map defects}  &
\makecell{WM-811K}   & 
\makecell{dense binary} & 
\makecell{binding; superposition} & 
\makecell{binarized centroids} & 
ANN; SVM \\ \hline

\end{tabular}
\end{table}

\begin{table}[t]
\tiny
\renewcommand{\arraystretch}{1.0}
\caption{HDC/VSA studies classifying automotive data
}
\centering
\label{tab:class:automotive}
\begin{tabular}{|c|c|c|c|c|c|c|}
\hline
Ref. & Task & Dataset & HV format & \makecell{Primitives used in  data transformation}  & Classifier & Baseline(s) \\ 
\hline

\makecell{ \cite{KleykoBrainlike2014} } & \makecell{Identification of vehicle type}  &  \makecell{From \cite{kleyko2015comparison}}     &  \makecell{dense  binary}   & \makecell{binding;  superposition}
 & \makecell{centroids}  & N/A\\ \hline

\makecell{ \cite{MirusReasoning2018} } & \makecell{Identification of   driving context}  &  \makecell{Own data}     &  \makecell{real-valued}   & \makecell{binding; superposition}
 & \makecell{spiking ANN}  & \makecell{ANN}\\ \hline

\makecell{ \cite{MirusSpatial2019};  \cite{MirusBehavior2019} } & \makecell{Prediction of  vehicle's trajectory}  &  \makecell{Own data; NGSIM US-101}     &  \makecell{real-valued}   & \makecell{binding;  superposition}
 & \makecell{LSTM}  & \makecell{LSTM  without HVs}\\ \hline

\makecell{ \cite{MirusBalanced2020} } & \makecell{Prediction of  vehicle's trajectory}  &  \makecell{From \cite{MirusBehavior2019}; NGSIM US-101}     &  \makecell{real-valued}   & \makecell{binding;  superposition}
 & \makecell{LSTM}  & \makecell{LSTM  without HVs}\\ \hline

\makecell{ \cite{MirusAbnormal2020} } & \makecell{Detection of abnormal  driving situations}  &  \makecell{From \cite{MirusBehavior2019}; NGSIM US-101}     &  \makecell{real-valued}   & \makecell{binding;  superposition}
 & \makecell{autoencoder  ANN}  & \makecell{N/A}\\ \hline

\cite{SchlegelMultivariate2021}  &
\makecell{Identification of driving style}   &  
\makecell{UAH-DriveSet}     & 
\makecell{complex-valued}     & 
\makecell{binding;  superposition; \\ fractional power encoding}     & 
\makecell{ANN; SNN; \\ SVM; $k$NN} &
LSTM without HVs  \\ \hline 

\makecell{\cite{WangAnomaly2021}} & 
\makecell{Detection of automotive \\ sensor attacks}  &
\makecell{AEGIS Big Data Project}   & 
\makecell{bipolar} & 
\makecell{binding; superposition} & 
\makecell{similarity between original \\ and reconstructed samples } & 
N/A \\ \hline

\end{tabular}
\end{table}

\begin{table}[H]
\tiny
\renewcommand{\arraystretch}{1.0}
\caption{HDC/VSA studies classifying behavioral signals}
\centering
\label{tab:class:beh:sig}
\begin{tabular}{|c|c|c|c|c|c|c|}
\hline
Ref. & Task & Dataset & HV format & \makecell{Primitives used in data transformation}  & Classifier & Baseline(s) \\ 
\hline
\cite{Rasanen2014} & Activity recognition  & Palantir   & bipolar & \makecell{binding   weighted superposition} 
 & centroids & N/A \\ \hline
\cite{KimActivity2018} & Activity recognition  & \makecell{ UCIHAR; PAMAP2;  EXTRA }     & bipolar & binding; superposition & \makecell{ binarized refined  centroids }  & binarized ANN \\ \hline
\cite{ChangEmotion2019} & Emotion Recognition  & AMIGOS     & \makecell{ sparse ternary;  binary }  & binding; superposition & binarized centroids  & XGBoost \\ \hline
\cite{Rasanen2015tr} & \makecell{Next GPS  location prediction}  & \makecell{Nokia  Lausanne }     &  sparse ternary   & weighted superposition & centroids  & \makecell{ mixed-order  Markov chain } \\ \hline
\cite{Rasanen2015tr} & \makecell{Next mobile  application prediction}  & \makecell{Nokia  Lausanne }     &  sparse ternary   & weighted superposition & centroids  & \makecell{ mixed-order  Markov chain } \\ \hline

\cite{Rasanen2015tr} & \makecell{Next singer  prediction}  & \makecell{Nokia  Lausanne }     &  sparse ternary   & weighted superposition & centroids  & \makecell{ mixed-order  Markov chain } \\ \hline

\cite{ImaniAdaptHD2019} & Activity recognition  &  UCIHAR    &  \makecell{dense  binary}   & \makecell{ binding;  superposition}
 & \makecell{binarized refined  centroids} &   \makecell{binarized  centroids} \\ \hline

\cite{ImaniMemoryCentric2020} & Activity recognition  & \makecell{ UCIHAR}      & binary & binding; superposition & \makecell{ multiple binarized  refined centroids  }  & $k$NN \\ \hline

\cite{BasaklarHypervector2021} & Activity recognition  &  \makecell{ From  \cite{BhatRecognition2020}}    &  bipolar   & \makecell{  superposition}
 & \makecell{centroids} &   \makecell{SVM} \\ \hline

\cite{BasaklarHypervector2021} & \makecell{Detection of  Parkinson's Disease}  &  \makecell{Parkinson's Disease  digital biomarker}    &  bipolar   & \makecell{  superposition}
 & \makecell{centroids} &   \makecell{SVM} \\ \hline

\cite{HernandezOnlineHD2021} & Activity recognition  & \makecell{ UCIHAR; PAMAP2}     & bipolar & binding; superposition & \makecell{ conditioned  centroids }  & ANN; SVM; AdaBoost \\ \hline

\cite{ZouManiHD2021} & Activity recognition  & \makecell{ UCIHAR}      & N/A & \makecell{trainable projection  matrix} & \makecell{ refined  centroids }  & ANN; SVM; AdaBoost \\ \hline

\cite{hsiao2021hyperdimensional} & Activity recognition  & \makecell{ UCIHAR}     & bipolar &  \makecell{ trainable projection matrix} & \makecell{ binarized refined  centroids }  & \makecell{binarized  centroids} \\ \hline

\cite{kazemi2021mimhd} & Activity recognition  & \makecell{ UCIHAR; PAMAP2}     & bipolar & binding; superposition & \makecell{ quantized   refined centroids }  & \makecell{ non-quantized   refined centroids } \\ \hline

\cite{ChangMulTaHDC2021} & Activity recognition  & \makecell{ UCIHAR}     & bipolar &  \makecell{binding; superposition} & \makecell{ binarized  centroids }  & \makecell{approach from \cite{ImaniSparseHD2019}} \\ \hline

\cite{ZhangRobustness2021} &  Activity recognition  & \makecell{ UCIHAR}     & bipolar & binding; superposition & \makecell{ quantized refined centroids }  & \makecell{N/A} \\ \hline

\cite{PoduvalStocHD2021} & Activity recognition  & \makecell{ UCIHAR; PAMAP2}     & bipolar & permutation; superposition & \makecell{ binarized refined  centroids }  & ANN; SVM; AdaBoost \\ \hline

\cite{YuUnderstandingHDC2022}  &
\makecell{Activity recognition}   &  
UCIHAR     & 
integer-valued &
binding; superposition  & 
\makecell{ discretized stochastic gradient descent}  & 
\makecell{approach from \cite{ImaniBinary2019}}  \\ \hline

\cite{HuRadarHumActiv2022} & 
Activity recognition  & 
\makecell{In-house based on LFMCW radar}     & integer-valued & 
binding; superposition & 
\makecell{refined centroids with masking}  & 
10 different methods  \\ \hline

\cite{MenonEmotionHD2021} & Emotion recognition  & \makecell{AMIGOS  DEAP}     & \makecell{dense  binary}  & \makecell{binding; permutation;  superposition} & binarized centroids  & XGBoost; SVM \\ \hline

\cite{MenonWearableCA2021} & Emotion recognition  & \makecell{AMIGOS}     & \makecell{dense  binary}  & \makecell{binding; permutation; superposition} & binarized centroids  & SVM \\ \hline

\end{tabular}
\end{table}

\begin{table}[H]
\tiny
\renewcommand{\arraystretch}{1.0}
\caption{HDC/VSA studies classifying EMG signals}
\centering
\label{tab:class:EMG}
\begin{tabular}{|c|c|c|c|c|c|c|}
\hline
Ref. & Task & Dataset & HV format & \makecell{Primitives used in  data transformation}  & Classifier & Baseline(s) \\ 
\hline

\makecell{\cite{RahimiBiosignal2016}} & \makecell{Hand gesture  recognition}  & \makecell{From \cite{BenattiEMG2014}}   & bipolar & \makecell{binding; permutation;  superposition}  
 & \makecell{conditioned  centroids} & SVM \\ \hline

\makecell{\cite{KleykoTradeoffs2018}} & \makecell{Hand gesture  recognition}  & \makecell{From \cite{BenattiEMG2014}}   & \makecell{sparse \& dense  binary} & \makecell{binding; permutation;  superposition}   & \makecell{conditioned  centroids} & \makecell{approach from  \cite{RahimiBiosignal2016}} \\ \hline

\makecell{\cite{MoinEMGFlexible2018}} & \makecell{Hand gesture  recognition}  & \makecell{Own data}   & bipolar & \makecell{binding; permutation;  superposition; scalar multiplication}  
 & \makecell{binarized  centroids} & N/A \\ \hline

\makecell{\cite{MontagnaUltraLow2018}} & \makecell{Hand gesture  recognition}  & \makecell{From \cite{BenattiEMG2014}}   & \makecell{dense  binary} & \makecell{binding; permutation;  superposition}  
 & \makecell{binarized  centroids} & SVM \\ \hline

\makecell{\cite{BenattiEMGGestures2019}} & \makecell{Hand gesture  recognition}  & \makecell{Own data}   & \makecell{dense  binary} & \makecell{binding; permutation;  superposition}  
 & \makecell{binarized centroids} & SVM \\ \hline

\makecell{\cite{MoinContractionGesture2019}} & \makecell{Hand gesture recognition with contraction levels}  & \makecell{Own data}   & bipolar & \makecell{binding; permutation;  superposition}  
 & \makecell{binarized  centroids} & N/A \\ \hline

\makecell{\cite{MoinWearable2021}} & \makecell{Hand gesture recognition}  & \makecell{Own data}   & bipolar & \makecell{binding; permutation; weighted superposition}  
 & \makecell{binarized  centroids} & N/A \\ \hline

 \makecell{\cite{ZhouGesture2021}} & \makecell{Adaptive hand gesture recognition}  & \makecell{Own data}   & bipolar & \makecell{binding; permutation;  weighted superposition}  
 & \makecell{context-aware  binarized  centroids} & \makecell{SVM; LDA} \\ \hline

 \makecell{\cite{ZhouIncremental2021}} & \makecell{Adaptive hand gesture recognition}  & \makecell{Own data}   & bipolar & \makecell{binding; permutation;  weighted superposition}  
 & \makecell{context-aware  binarized  centroids} & \makecell{SVM; LDA} \\ \hline

\makecell{\cite{EggimannConfigurableHD2021}} & \makecell{Hand gesture  recognition}  & \makecell{From  \cite{MoinEMGFlexible2018}}   & \makecell{dense  binary} & \makecell{binding; permutation;  superposition}  
 & \makecell{binarized  centroids} & \makecell{approach from  \cite{MoinEMGFlexible2018}} \\ \hline

\makecell{\cite{KarunaratneSTsignals2021}} & \makecell{Hand gesture  recognition}  & \makecell{From  \cite{RahimiBiosignal2016}}   & \makecell{dense  binary} & \makecell{binding; permutation;  superposition} & \makecell{binarized  centroids} & \makecell{approach from  \cite{MontagnaUltraLow2018}} \\ \hline

\end{tabular}
\end{table}

\paragraph{Automotive data}

Table~\ref{tab:class:automotive}\footnote{
It should be noted that works~\cite{MirusBehavior2019, MirusBalanced2020} were, strictly speaking, solving the regression problems while~\cite{MirusAbnormal2020, WangAnomaly2021} were concerned with anomaly detection. These studies are listed in this section for the sake of covering the applications within the automotive data.
} 
presents studies where HDC/VSA was used with automotive data, 
mainly in autonomous driving scenarios.

\paragraph{Behavioral signals}

Studies that used behavioral signals are summarized in Table~\ref{tab:class:beh:sig}. 
One of the most common applications was activity recognition, but other tasks were considered as well (see the table).

\paragraph{Biomedical data}

Currently explored applications of HDC/VSA on biomedical data can be categorized into five types of data: electromyography (EMG) signals, electroencephalography (EEG) signals, cardiotocography (CTG) signals, DNA sequences, and surface enhanced laser desorption/ionization time-of-flight (SELDI-TOF) mass spectrometry.
Most of the works so far have been conducted on EMG and EEG signals. In fact, there is a study~\cite{RahimiBiosignal2019}, which provides an in-depth coverage of applying HDC/VSA to these modalities, so please refer to this article for a detailed overview of the area.

\noindent{\textbf{EMG signals.}} HDC/VSA was applied to EMG signals for the task of hand gesture recognition. 
This was done for several different transformations of data into HVs and on different datasets. Refer to Table~\ref{tab:class:EMG} for the summary.

\noindent{\textbf{EEG and iEEG signals.}}
EEG and iEEG signals were used with HDC/VSA for human-machine interfaces and epileptic seizure detection. 
These efforts are overviewed in Table~\ref{tab:class:EEG}. 
It is worth mentioning systematization efforts in~\cite{PaleHDEpileptic2021,UnaSelection2022}, which reported an assessment of several HDC/VSA models and transformations used for epileptic seizure detection.
Another recent work~\cite{SchindlerPrimerHDiEEG2021} provides a tutorial on applying HDC/VSA for iEEG seizure detection.

\begin{table}[t]
\tiny
\renewcommand{\arraystretch}{1.0}
\caption{HDC/VSA studies classifying EEG signals}
\centering
\label{tab:class:EEG}
\begin{tabular}{|c|c|c|c|c|c|c|}
\hline
Ref. & Task & Dataset & HV format & \makecell{Primitives used in  data transformation}  & Classifier & Baseline(s) \\ 
\hline
\makecell{\cite{RahimiBrainComputer2017};  \cite{RahimiEEG2017}} & \makecell{Subject's intentions  recognition}  & \makecell{Monitoring error- related potentials}   & bipolar & \makecell{binding; superposition; permutation}
 & \makecell{conditioned  centroids} & \makecell{Gaussian  classifier} \\ \hline
 
 \cite{BasaklarHypervector2021} & \makecell{Subject's intentions  recognition}  &  \makecell{Monitoring error- related potentials}    &  bipolar   & \makecell{  superposition}
 & \makecell{centroids} &   \makecell{SVM} \\ \hline 
 
\cite{RahimiBiosignal2019} & \makecell{Multiclass subject's intentions recognition}  &  \makecell{4-class EEG motor imagery signals}    &  bipolar   & \makecell{binding; superposition}
 & \makecell{binarized centroids} &   \makecell{CNN} \\ \hline 
 
\cite{HerscheExpEmbedding2018} & \makecell{Multiclass subject's intentions recognition}  &  \makecell{4-class and 3-class EEG motor imagery}    &  \makecell{dense binary}   & \makecell{random and trainable \\ projection; superposition}
 & \makecell{multiple centroids } &   \makecell{SVM} \\ \hline

\cite{BurrelloSeizure2018} & \makecell{Epileptic seizure  detection}  & \makecell{Short-term  SWEC-ETHZ iEEG}   & \makecell{dense  binary} & binding; superposition
 & \makecell{binarized  centroids} & ANN\\ \hline

\cite{EnsembleBurrello2021} & 
\makecell{Epileptic seizure detection}  & 
\makecell{Short-term  SWEC-ETHZ iEEG}   & 
\makecell{dense binary} & 
\makecell{binding; superposition\\ each operating on \\  a different feature set} & 
\makecell{Ensemble of binarized \\ centroids combined \\ via a linear layer} & 
ANN; SVM; CNN\\ \hline

 \cite{BurrelloLBPSeizure2020} & \makecell{Epileptic seizure  detection \& \\ Identification of Ictogenic Brain Regions}  & \makecell{Short-term \\ SWEC-ETHZ iEEG}   & \makecell{dense\\  binary} & binding; superposition
 & \makecell{binarized \\ centroids} & \makecell{ANN, LSTM,\\SVM, RF}\\ \hline

\cite{BurrelloSeizure2019} & \makecell{Epileptic seizure  detection}  & \makecell{Long-term  SWEC-ETHZ iEEG}  & \makecell{dense  binary} & binding; superposition
 & \makecell{binarized  centroids} & \makecell{LSTM, CNN, SVM}\\ \hline

\cite{AsgarinejadHDSeizures2020} & \makecell{Epileptic seizure  detection}  & CHB-MIT Scalp EEG   & bipolar & binding; superposition
 & centroids & CNN\\ \hline 

\cite{UnaMulti-Centroid2022} & 
\makecell{Epileptic seizure  detection}  & 
CHB-MIT Scalp EEG   & 
bipolar & binding; superposition
 &  \makecell{multi-centroids \\ for sub-classes} &  \makecell{single centroid \\ HDC/VSA solution}
\\ \hline 
 
\cite{GeSeizure2021} & \makecell{Epileptic seizure  detection}  &  \makecell{UPenn and Mayo  Clinic's Seizure Detection}    &  \makecell{dense  binary}   & \makecell{ binding;  superposition}
 & \makecell{binarized  centroids} &   \makecell{SVM} \\ \hline

\cite{PaleHDEpileptic2021} & \makecell{Epileptic seizure  detection}  &  \makecell{Short-term  SWEC-ETHZ iEEG;  CHB-MIT Scalp EEG }    &  \makecell{dense  binary}   & \makecell{ binding;  superposition}
 & \makecell{binarized centroids} &   \makecell{SVM} \\ \hline 
 
 \cite{GeSeizure2022}  &
\makecell{Epileptic seizure  detection}   &  
\makecell{UPenn and Mayo  Clinic's Seizure Detection}     & 
dense  binary &
binding; superposition  & 
\makecell{binarized centroids}  & 
\makecell{N/A}  \\ \hline

\end{tabular}
\end{table}

\noindent{\textbf{CTG signals}}. 
So far, there is only one work~\cite{BasaklarHypervector2021} where CTG signals were used. 
It is summarized in Table~\ref{tab:class:Cardiotocography}.

\begin{table}[t]
\tiny
\renewcommand{\arraystretch}{1.0}
\caption{HDC/VSA studies classifying Cardiotocography signals}
\centering
\label{tab:class:Cardiotocography}
\begin{tabular}{|c|c|c|c|c|c|c|}
\hline
Ref. & Task & Dataset & HV format & \makecell{Primitives used in  data transformation}  & Classifier & Baseline(s) \\ 
\hline

\cite{BasaklarHypervector2021} & \makecell{Classification of  fetal state}  &  \makecell{Cardiotocography}    &  bipolar   & \makecell{  superposition}
 & \makecell{centroids} &   \makecell{SVM} \\ \hline

\end{tabular}
\end{table}

\begin{table}[H]
\tiny
\renewcommand{\arraystretch}{1.0}
\caption{HDC/VSA studies classifying DNA sequences}
\centering
\label{tab:class:beh:dna}
\begin{tabular}{|c|c|c|c|c|c|c|}
\hline
Ref. & Task & Dataset & HV format & \makecell{Primitives used in  data transformation}  & Classifier & Baseline(s) \\ 
\hline
\makecell{\cite{ImaniDNA2018}} & \makecell{DNA classification}  & \makecell{Empirical;  Molecular Biology}   & \makecell{dense  binary} & \makecell{binding; permutation;  superposition} 
 & \makecell{binarized centroids}  & \makecell{$k$NN; SVM} \\ \hline

\makecell{ \cite{CumboDNACancer2020} } & \makecell{Detection of tumor}  &  \makecell{
BRCA; KIRP; THCA}     &  \makecell{bipolar}   & \makecell{permutation;  superposition}
 & \makecell{refined  centroids}  & SVM \\ \hline
 
 \cite{RachkovskijEquivariant2021} &
\makecell{Recognition of of splice junctions}   &  
\makecell{Splice-junction Gene Sequences}     & 
sparse  binary &
permutation; superposition  & 
\makecell{centroids, SVM, $k$NN}  & 
\makecell{CNN, $k$NN}  \\ \hline

 \cite{RachkovskijEquivariant2021}  &
\makecell{Prediction of protein's secondary structure}   &  
\makecell{Protein Secondary Structure}     & 
sparse  binary &
permutation; superposition  & 
\makecell{centroids, SVM, $k$NN}  & 
\makecell{ANN}  \\ \hline

\end{tabular}
\end{table}

\begin{table}[H]
\tiny
\renewcommand{\arraystretch}{1.0}
\caption{HDC/VSA studies classifying the sensitivity of glioma to chemotherapy using SELDI-TOF}
\centering
\label{tab:class:beh:spectrometry}
\begin{tabular}{|c|c|c|c|c|c|c|}
\hline
Ref. & Task & Dataset & HV format & \makecell{Primitives used in  data transformation}  & Classifier & Baseline(s) \\ 
\hline

 \makecell{ \cite{RachkovskijGliomas2010}} & \makecell{Glioma sensitivity classification}  &  \makecell{Cancer-Glioma} &  \makecell{binary}   & \makecell{RSC}
 & \makecell{SVM}  &  \makecell {MLP; Probabilistic NN;  Associative memory}\\ \hline 

\end{tabular}
\end{table}

\begin{table}[H]
\tiny
\renewcommand{\arraystretch}{1.0}
\caption{HDC/VSA studies classifying multi-modal signals}
\centering
\label{tab:class:beh:multi}
\begin{tabular}{|c|c|c|c|c|c|c|}
\hline
Ref. & Task & Dataset & HV format & \makecell{Primitives used in  data transformation}  & Classifier & Baseline(s) \\ 
\hline

\cite{ChangEmotion2019} & Emotion Recognition  & AMIGOS     & \makecell{ sparse ternary; binary }  & binding; superposition & binarized centroids  & XGBoost \\ \hline

\cite{MenonEmotionHD2021} & Emotion recognition  & \makecell{AMIGOS  DEAP}     & \makecell{dense   binary}  & \makecell{binding; permutation;  superposition} & binarized centroids  & XGBoost; SVM \\ \hline

\cite{MenonWearableCA2021} & Emotion recognition  & \makecell{AMIGOS}     & \makecell{dense  binary}  & \makecell{binding; permutation; superposition} & binarized centroids  & SVM \\ \hline

 \makecell{ \cite{WatkinsonSeptic2021} } & \makecell{Septic shock  detection}  &  \makecell{eICU }     &  \makecell{dense  binary}   & \makecell{permutation; binding;  superposition}
 & \makecell{nearest  neighbor}  & N/A \\ \hline 

\end{tabular}
\end{table}

\noindent{\textbf{DNA sequences.}}
Table~\ref{tab:class:beh:dna} presents two studies that used DNA sequences in classification tasks.

\noindent{\textbf{SELDI-TOF mass spectrometry.}} Table~\ref{tab:class:beh:spectrometry} summarizes a study that used SELDI-TOF mass spectrometry for classifying sensitivity of glioma to chemotherapy.

\noindent{\textbf{Multi-modal signals.}} Studies involding multi-modal signals are summarized in Table~\ref{tab:class:beh:multi}.

\subsubsection{Classification of images or their properties}
\label{sec:app:class:images}

Table~\ref{tab:class:images} provides an overview of the efforts involving images. 
Since using raw pixels directly would rarely result in a good performance, HVs were produced either from features extracted from images or using HVs obtained from neural networks (see Section~\ref{PartI-sec:2Dimages:neural:nets} in~\cite{KleykoSurveyVSA2021Part1}), which took images as an input.

\begin{table}[t]
\tiny
\renewcommand{\arraystretch}{1.0}
\caption{HDC/VSA studies classifying visual images and their properties}
\centering
\label{tab:class:images}
\begin{tabular}{|c|c|c|c|c|c|c|}
\hline
Ref. & Task & Dataset & HV format & \makecell{Primitives used in \\ data transformation}  & Classifier & Baseline(s) \\ 
\hline
\makecell{\cite{RachkovskijTexture1991}} & \makecell{Texture classification}  & \makecell{Own data}   & \makecell{sparse   binary} & \makecell{binding;  superposition} 
 & \makecell{perceptron-like  algorithm}  & \makecell{N/A} \\ \hline

    \cite{KussulPermutation2006} & \makecell{Micro-object\\
shape recognition}  &  \makecell{Own data}    & \makecell{sparse \\  binary}  & \makecell{RLD HV permutations;\\
 superposition}
 & \makecell{large margin \\ perceptron} & \makecell{N/A} \\ \hline

 \makecell{ \cite{KleykoModality2017} } & \makecell{Modality classification}  &  \makecell{
IMAGE  CLEF2012}     &  \makecell{dense  binary}   & \makecell{cellular automata;  superposition}
 & \makecell{centroids}  & \makecell{SVM; $k$NN} \\ \hline  

\makecell{\cite{ParhiGenClass2021}} & 
\makecell{Biological gender classification}  &  \makecell{fMRI from HCP}     &  
\makecell{bipolar}   & 
\makecell{binding; superposition} & 
\makecell{non-quantized refined centroids}  & \makecell{Random Forest; PCA; etc.} \\ \hline

 \makecell{ \cite{NeubertRobotics2019} } & \makecell{Visual place  recognition}  &  \makecell{
Nordland dataset}     &  \makecell{bipolar}   & \makecell{permutation; binding;  superposition}
 & \makecell{centroids}  & SeqSLAM \\ \hline 

 \makecell{ \cite{NeubertPlaceRecognition2021} } & \makecell{Visual place  recognition}  &  \makecell{
OxfordRobotCar; StLucia; \\ CMU Visual Localization }     &  \makecell{complex-valued}   & \makecell{permutation; binding; \\ superposition}
 & \makecell{centroids}  & \makecell{AlexNet; HybridNet; \\ NetVLAD; DenseVLAD}  \\ \hline 

 \makecell{ \cite{MitrokhinSensorimotor2019} } & \makecell{Ego-motion estimation}  &  \makecell{
MVSEC}     &  \makecell{dense  binary}  & \makecell{permutation; binding;  superposition}
 & \makecell{centroids}  & N/A  \\ \hline 

 \makecell{ \cite{HerscheDVSCDT2020} } & \makecell{Ego-motion estimation}  &  \makecell{
MVSEC}     &  \makecell{sparse\\  binary}  & \makecell{random sparse projection;  CDT}
 & \makecell{incremental \\centroids}  & \makecell{dense binary/integer;\\ ANN; various regressions}  \\ \hline 

 \makecell{ \cite{WatkinsonPneumonia2021} } & \makecell{Detection of pneumonia}  &  \makecell{SARS-CoV-2 CT-scan \cite{he2020sample} \& \cite{rahimzadeh2021fully} }     &  \makecell{bipolar}   & \makecell{binding;  superposition}
 & \makecell{binarized centroids}  & ANN \\ \hline
 
\makecell{ \cite{GallantPositional2016} } & \makecell{Object classification}  &  \makecell{CIFAR-10;  Artificial Face}     &  \makecell{dense  binary}   & \makecell{positional binding;  superposition}
 & \makecell{ridge  regression}  & CNN \\ \hline 

 \makecell{ \cite{MitrokhinCNN2020} } & \makecell{Object classification}  &  \makecell{
CIFAR-10;  NUSWIDE\_81}     &  \makecell{dense  binary}   & \makecell{permutation; binding;  superposition}
 & \makecell{centroids}  & N/A \\ \hline 

\cite{YuUnderstandingHDC2022}  &
\makecell{Object classification}   &  
Fashion-MNIST     & 
integer-valued &
binding; superposition  & 
\makecell{ discretized stochastic  gradient descent}  & 
\makecell{approach from \cite{ImaniBinary2019}}  \\ \hline  \hline

  \cite{KussulLiRA2004} & \makecell{Character recognition}  &  \makecell{MNIST}    & \makecell{sparse   binary}  & \makecell{LIRA}
 & \makecell{large margin  perceptron} & \makecell{conventional classifiers} \\ \hline
 
  \cite{KussulPermutation2006} & \makecell{Character recognition}  &  \makecell{MNIST}    & \makecell{sparse   binary}  & \makecell{RLD HV permutations;
 superposition}
 & \makecell{large margin perceptron} & \makecell{conventional classifiers} \\ \hline
 
 \cite{RachkovskijClassifiers2007} & \makecell{Character recognition}  &  \makecell{MNIST}    & \makecell{sparse   binary}  & \makecell{LIRA HV; superposition}
 & \makecell{large margin  perceptron} & \makecell{feature  selection} \\ \hline
 
 \makecell{ \cite{ManabatCharacter2019} } & \makecell{Character recognition}  &  \makecell{
MNIST}     &  \makecell{dense  binary}   & \makecell{permutation; binding; superposition}
 & \makecell{centroids}  & N/A \\ \hline 
 
  \makecell{ \cite{KarvonenFPGA_CA_HD2019} } & \makecell{Character recognition}  &  \makecell{
MNIST}     &  \makecell{N/A}   & \makecell{cellular automata-based}
 & \makecell{random centroids}  & Naïve Bayes \\ \hline

\cite{ChuangHDTradeoffs2020} & \makecell{Character recognition}  &  \makecell{MNIST} & binary &  \makecell{binding; superposition} & \makecell{ non-quantized   refined centroids }   & \makecell{refined  centroids} \\ \hline

\cite{HernandezOnlineHD2021} & \makecell{Character recognition}  &  \makecell{MNIST}     & bipolar & binding; superposition & \makecell{ conditioned  centroids }  & ANN; SVM; AdaBoost \\ \hline

\cite{kazemi2021mimhd} &  \makecell{Character recognition}  &  \makecell{MNIST}      & bipolar & binding; superposition & \makecell{ quantized   refined centroids }  & \makecell{ non-quantized   refined centroids } \\ \hline

\cite{ZouManiHD2021} & \makecell{Character recognition}  &  \makecell{MNIST} & N/A & \makecell{trainable projection  matrix} & \makecell{ refined  centroids }  & ANN; SVM; AdaBoost \\ \hline

\cite{ChangMulTaHDC2021} & \makecell{Character recognition}  &  \makecell{MNIST} & bipolar &  \makecell{binding; superposition} & \makecell{ binarized  centroids }  & \makecell{Approach from \cite{ImaniSparseHD2019}} \\ \hline

\cite{PoduvalStocHD2021} & \makecell{Character recognition}  &  \makecell{MNIST}     & bipolar & permutation; superposition & \makecell{ binarized refined  centroids }  & ANN; SVM; AdaBoost \\ \hline

\cite{YuUnderstandingHDC2022}  &
\makecell{Character recognition}  &  
MNIST     & 
integer-valued &
binding; superposition  & 
\makecell{ discretized stochastic gradient descent}  & 
\makecell{approach from \cite{ImaniBinary2019}}  \\ \hline
\hline

 \makecell{\cite{KarunaratneHDAugmented2021}} & \makecell{Few-shot character recognition}  &  \makecell{Omniglot}     &  \makecell{dense binary}   & \makecell{5-layer CNNs}
 & \makecell{weighted $k$NN }  & various CNNs \\ \hline

\makecell{\cite{KleykoAugmented2022}} & 
\makecell{Few-shot character recognition}  &  
\makecell{Omniglot} &  
\makecell{dense binary} & 
\makecell{5-layer CNNs} & 
\makecell{Outer product-based \\associative memory}  & 
approach from~\cite{KarunaratneHDAugmented2021} \\ \hline 
 \hline

\makecell{\cite{HerscheContinualLearn2022}} & \makecell{Few-shot continual learning \\for image classification}  &  
\makecell{CIFAR-100; miniImageNet;\\ Omniglot} &
\makecell{real-valued} & 
\makecell{Pre-trained CNN and \\  a retrainable linear layer} & \makecell{(bipolar) centroids;\\ loss-optimized nudged centroids}  & 
various deep ANNs \\ \hline \hline
 
   \cite{KussulPermutation2006} & \makecell{Face recognition}  &  \makecell{ORL}    & \makecell{sparse   binary}  & \makecell{RLD HV permutations; superposition}
 & \makecell{large margin  perceptron} & \makecell{conventional classifiers} \\ \hline

\cite{ImaniMemoryCentric2020} & \makecell{Face recognition}  &  \makecell{FACE}     & binary & binding; superposition & \makecell{ multiple binarized refined centroids  }  & $k$NN \\ \hline

\cite{HernandezOnlineHD2021} & \makecell{Face recognition}  &  \makecell{FACE}     & bipolar & binding; superposition & \makecell{ conditioned  centroids }  & ANN; SVM; AdaBoost \\ \hline

\cite{Curtidor21} &  \makecell{Face recognition}  &  \makecell{ORL; FRAV3D; FEI}      & \makecell{sparse  binary}  &  \makecell{RLD; permutation;
 superposition}
 & \makecell{large margin perceptron} & \makecell{ SVM; Iterative Closest Point } \\ \hline

\cite{kazemi2021mimhd} &  \makecell{Face recognition}  &  \makecell{FACE}      & bipolar & binding; superposition & \makecell{ quantized  refined centroids }  & \makecell{ non-quantized   refined centroids } \\ \hline

\cite{PoduvalStocHD2021} & \makecell{Face recognition}  &  \makecell{FACE}     & bipolar & permutation; superposition & \makecell{ binarized refined  centroids }  & ANN; SVM; AdaBoost \\ \hline

\end{tabular}
\end{table}

\subsubsection{Classification of structured data}
\label{sec:app:class:structured}

\begin{table}[t]
\tiny
\renewcommand{\arraystretch}{1.0}
\caption{HDC/VSA studies classifying structured data}
\centering
\label{tab:class:structured}
\begin{tabular}{|c|c|c|c|c|c|c|}
\hline
Ref. & Task & Dataset & HV format & \makecell{Primitives used in  data transformation}  & Classifier & Baseline(s) \\ 
\hline

\makecell{\cite{SlipchenkoHierarchical2005}} & 
\makecell{Prediction of chemical \\ compound properties}  &  
\makecell{INTAS00-397} &  
\makecell{sparse binary} & 
\makecell{binding; superposition;} & 
\makecell{SVM}  & 
DISCOVERY; ANALOGY \\ \hline

\makecell{\cite{MaDrugDiscovery2021}} & 
\makecell{Drug discovery}  &  
\makecell{Clintox; BBBP; SIDER} &  
\makecell{bipolar} & 
\makecell{permutation; superposition;} & 
\makecell{refined centroids}  & 
Logistic Regression; SVM; Random Forest; etc. \\ \hline

\end{tabular}
\end{table}

Classification of structured data can be tricky with conventional machine learning algorithms since local representations of structured data might not be convenient to work when using vector classifiers, especially when the data involve some sorts of hierarchies. 
HDC/VSA should be well suited for structured data since they allow representing various structures (including hierarchies) as HVs. 
To the best of our knowledge, however, the number of such studies is very limited (see Table~\ref{tab:class:structured}). 
One such study in \cite{SlipchenkoHierarchical2005} used SBDR to predict the properties of chemical compounds and provided the state-of-the-art performance.  
A more recent example was demonstrated in~\cite{MaDrugDiscovery2021}, where 2D molecular structures were transformed into HVs that were used to construct classifiers for drug discovery problems. The approach outperformed the baseline
methods on a collection of $30$ tasks. 
Finally, in~\cite{NunesGraphHD2022} it was proposed to classify graphs using HDC/VSA.   
A graph was represented as a superposition of HVs corresponding to vertices and edges. 
The proposed approach was evaluated on six graph classification datasets; when compared to the baseline approaches, it demonstrated comparable accuracy on four datasets and much shorter training time on all of them.

\section{Cognitive computing and architectures}
\label{sec:cog}

In this section, we overview the use of HDC/VSA in cognitive computing (Section~\ref{sec:cog:comp}) and cognitive architectures (Section~\ref{sec:cog:arch}). 
Note that, strictly speaking, cognitive computing as well as cognitive architectures can also be considered to be application areas but we decided to separate them into a distinct section due to the rather different nature of tasks being pursued.

\subsection{Cognitive computing}
\label{sec:cog:comp}

\subsubsection{Holistic transformations}
\label{sec:database:holistic}

\paragraph{Holistic transformations in database processing with HVs}
\label{sec:holistic:proc}

In this section, we consider simple examples of HDC/VSA processing of a small database. 
This can be treated as analogical reasoning  (though analogy researchers might disagree) in the form of query answering using simple analogs (we refer to them as records) without explicitly taking into account the constraints on analogical reasoning mentioned in the following sections. 

The work \cite{KanervaFully1997} introduced the transformation of database records into HVs as an alternative to their symbolic representations. 
Each record is a set of role-filler bindings and is represented by HVs using transformations for role-filler bindings (Section~\ref{PartI-sec:key:value} in~\cite{KleykoSurveyVSA2021Part1}) and sets (Section~\ref{PartI-sec:sets} in~\cite{KleykoSurveyVSA2021Part1}). We might be interested in querying the whole base for some property or in processing a pair of records. 
For example, knowing a filler of one of the roles in one record (the role is not known), can enable to get the filler of that role in another record. In \cite{KanervaFully1997}, the records were persons with attributes (i.e., roles) such as ``name'', ``gender'', and ``age'' (see Table \ref{tab:persons}).

Several ways were proposed to use HDC/VSA operations for querying the records by HVs. An example query to the database could be ``What is the age of Lee who is a female?''. The correct record for this query will be $\mathbf{LF6}$ and the answer is $66$. Depending on the prior knowledge, different cases were considered. 
All the cases below assume that there is an item memory with the HVs of all the roles, fillers, and database records.

\textit{Case} 1. We know the role-filler bindings name:Lee, gender:female, and only the role ``age'' from the third role-filler binding whose filler we want to find.
\textit{Solution} 1. The query is represented as $\mathbf{name} \circ \mathbf{Lee}$ + $\mathbf{gender}\circ \mathbf{female}$ . The base memory will return $\mathbf{LF6}$ as the closest match using the similarity of HVs. Unbinding $\mathbf{age} \oslash \mathbf{LF6}$ results in the noisy version of HV for $\mathbf{66}$, and the clean-up procedure returns the value associated with the nearest HV in the item memory, i.e., $\mathbf{66}$, which is the answer.

\begin{table}[t]
\small
\renewcommand{\arraystretch}{1.0}
\caption{
Example database records (adapted from~\cite{KanervaFully1997}.)
}
\label{tab:persons}
\centering
\begin{tabular}{|c|c|c|c|}
\hline
Record & Name & Gender & Age  \\
\hline
$\mathbf{PF3}$ & Pat & female & 33 \\ 
\hline
$\mathbf{PM6}$ & Pat & male & 66 \\
\hline
$\mathbf{LF6}$ & Lee & female & 66 \\
\hline
$\mathbf{LM3}$ & Lee & male & 33 \\
\hline
$\mathbf{LM6}$ & Lee & male & 66\\

\hline
\end{tabular}
\end{table}

\textit{Case} 2. The following HVs are available: record $\mathbf{PM6}$ and the fillers $\mathbf{Pat}$, $\mathbf{male}$, and $\mathbf{Lee}$,  as well as the roles $\mathbf{female}$ and $\mathbf{age}$. 
\textit{Solution} 2a. First, we find $\mathbf{name}$ by the clean-up procedure on $\mathbf{Pat} \oslash \mathbf{PM6}$, and $\mathbf{gender}$ by the clean-up procedure on $\mathbf{male} \oslash \mathbf{PM6}$. Then we apply the previous solution (\textit{Solution} 1). \textit{Solution} 2b. This solution\footnote{
This solution relies on knowing that there is only one Lee \& female record in the database, so it should not be considered as a sensible database operation but as an example demonstrating substitution transformations.
} uses the correspondences $\mathbf{Pat} \leftrightarrow \mathbf{Lee}$ and $\mathbf{male} \leftrightarrow \mathbf{female}$ by forming the transformation HV $\mathbf{T} = \mathbf{Pat}\circ \mathbf{Lee} + \mathbf{male}\circ \mathbf{female} $. 
The transformation $\mathbf{T} \oslash \mathbf{PM6}$ returns approximate $\mathbf{LF6}$ (see \cite{KanervaFully1997} for a detailed explanation), and its clean-up provides exact $\mathbf{LF6}$. Then, as above, $\mathbf{age} \oslash \mathbf{LF6}$ after the clean-up procedure returns $\mathbf{66}$. 
Note that such a transformation is intended for HDC/VSA models where the binding operation is self-inverse (e.g., BSC).  

\textit{Case} 3. Only $\mathbf{33}$ and $\mathbf{PF3}$ are known, and the task is to find an analog of  $\mathbf{33}$ in $\mathbf{LF6}$, that is, $\mathbf{66}$. A trivial solution would be to get $\mathbf{age}$  as the result of the clean-up procedure for $\mathbf{33} \oslash \mathbf{PF3}$, and then get $\mathbf{33}$ by $\mathbf{age} \oslash \mathbf{PF3}$ and the clean-up procedure. But there is a more interesting way, which can be considered as an analogical solution. 
\textit{Solution} 3. One step solution is $ \mathbf{LF6} \oslash (\mathbf{33}\oslash \mathbf{PF3})$. 
This exemplifies a possibility of processing without the intermediate use of a clean-up procedure. 

In some HDC/VSA models (e.g., BSC), however, the answer for this solution will be ambiguous, being equally similar to $\mathbf{33}$ and $\mathbf{66}$. This is due to the self-inverse property of the binding operation in BSC. Note that both $\mathbf{LF6}$ and $\mathbf{PF3}$ include $\mathbf{gender}\circ \mathbf{female}$ as the part of their records. Unbinding $\mathbf{33}$ with $\mathbf{PF3}$ creates $ \mathbf{33} \circ  \mathbf{gender}\circ \mathbf{female}$ (since $\oslash=\circ$) among other bindings. When unbinding the result with $\mathbf{LF6}$, the HV $\mathbf{gender}\circ \mathbf{female}$ will cancel out, thus releasing $\mathbf{33}$, which will interfere with the correct answer, which is $\mathbf{66}$. 
This effect would not appear if the records would had different fillers in their role-filler bindings. For example, if instead of $\mathbf{LF6}$ we consider $\mathbf{LM6}$, then $\mathbf{33} \circ \mathbf{PF3} \circ \mathbf{LM6}$ produces the correct answer. 

In models with self-inverse binding, the result of $\mathbf{PF3} \oslash \mathbf{LM6}$ can be seen as an interpretation of $\mathbf{PF3}$ in terms of $\mathbf{LM6}$ or vice versa, because the result of this operation is  
$\mathbf{Pat} \circ \mathbf{Lee} + \mathbf{male} \circ \mathbf{female} + \mathbf{33} \circ \mathbf{66} + \mathrm{noise} $ (since $\oslash=\circ$). 
This allows answering queries of the form ``which filler in $\mathbf{PF3}$ plays the same role as (something) in $\mathbf{LM6}$?'' by unbinding $\mathbf{PF3} \oslash \mathbf{LM6}$ with the required filler HV, resulting in the noisy answer HV.

Note that Solution 1 resembles the standard processing of records in databases. We first identify the record and then check the value of the role of interest. \textit{Solutions} 2b and 3 are examples of a different type of computing sometimes called ``holistic mapping'' \cite{KanervaFully1997} or ``transformation without decomposition'' \cite{PlateStructure1997}. We call it 
``holistic transformation'' in order not to confuse it with the transformation of input data into HVs or with mapping in analogical reasoning. This holistic transformation of HVs is commonly illustrated by an example well-known under the name ``Dollar of Mexico'' \cite{KanervaAnalogy1998, KanervaLarge2001, KanervaDollar2010}. 

In essence, the ``Dollar of Mexico'' show-case solves some simple ``proportional analogies'' of the form A:B~::~C:D as (i.e., United States:Mexico~::~Dollar:?). 
These analogies are also known to be solvable by addition and subtraction of the ``neural'' word embeddings of the corresponding concepts~\cite{MikolovWord2vec2013, pennington2014glove}.
Following a similar approach, the authors in~\cite{PaulladaRetrieval2020} proposed to improve the results by training shallow neural networks using a dependency path of relations between terms in sentences.

It should be noted that there is no direct analog to this kind of processing by holistic transformation 
(using geometric properties of the representational space) in the conventional symbol manipulation \cite{PlateStructure1997}. The holistic transformation of HVs can be seen as a parallel alternative to the conventional sequential search.

\paragraph{Learning holistic transformations from examples}
\label{sec:transformation}

Learning systematic transformations from examples was investigated in~\cite{NeumannTransformation2000, NeumannTransformation2002} for HRR. Previously, this capability was shown in~\cite{PlateStructure1997} only for manually constructing the so-called transformation HV. In~\cite{NeumannTransformation2000, NeumannTransformation2002}, the transformation HV was obtained from several training pairs of HVs. One of the proposed approaches to obtaining the transformation HV was to use the gradient descent by iterating through all examples until the optimization converged. The experiments demonstrated that the learned transformation HVs were able to generalize to previously unseen compositional structures with novel elements. A high level of systematicity was indicated by the ability of transformation HVs  to generalize to novel elements in structures of a complexity higher than the structures provided as training examples. The capability of BSC to learn holistic transformations was also presented in \cite{KanervaLearning2000, KanervaLarge2001}.
However, the disadvantage of such holistic transformations is their bidirectionality, which is due to the fact that the unbinding operation in BSC is equivalent to the binding operation.  This complication can be resolved by using either the permutation or an additional associative memory as the binding operation as proposed in \cite{EmruliAnalogical2014}.

The holistic transformation of the kind considered above was used to associate, e.g., sensory and motor/action data via binding HVs. For example, in \cite{KleykoBees2015, KleykoFlyBee2015}, BSC was applied to form a scene representation in the experiments with honey bees. It was shown to mimic learning in honey bees using the transformation as in \cite{KanervaLearning2000, KanervaLarge2001}.   
In~\cite{EmruliInteroperability2015},  an HDC/VSA-based approach for learning behaviors, based on observing and associating sensory and motor/action data represented by HVs, was proposed. 
A potential shortcoming of the approaches to learning holistic transformations presented above is that the objects/relations are assumed to be dissimilar to each other. The learning might not work as expected given that there is some similarity structure between the objects/relations used as training examples. This direction deserves further investigation.

\subsubsection{Analogical reasoning}
\label{sec:analogical:reasoning}

We begin with a brief introduction to analogical reasoning by summarizing basic analogical processes as well as some of their properties. 
Two final sections discuss applications of HDC/VSA to modeling analogical retrieval and mapping.

\paragraph{Basics of analogical reasoning modeling}

Modeling analogical reasoning in humans is an important frontier for Artificial Intelligence. Because it allows analogical problem solving as well as even more universal cognitive processes taking the problem structure into account and applying knowledge acquired in different domains. 
Analogical reasoning theories~\cite{GentnerStructure1983, KokinovAnalogy2003,GentnerAnalogical2010, GentnerAnalogical2012, GentnerAnalogical2017} usually consider and model the following basic processes: description; retrieval (also known as access or search); mapping; and inference. 

Any model for analogical reasoning usually works with analogical episodes (or simply analogs).
The description process concerns the representation of analogical episodes.
The analogical episodes are usually modeled as systems of hierarchical relations (predicates), consisting of elements in the form of entities (objects) and relations of various hierarchical levels. Entities belong to some subject domain (e.g., the sun, planet, etc.) and are described by attributes (features, properties), which in essence are relations with a single argument (e.g., mass, temperature, etc.). Relations  (e.g., attract, more, cause, etc.) define relations between the elements of analogical episodes. Arguments of relations may be objects, attributes, and other relations. It is assumed that a collection of base (source) analogical episodes is stored in a memory. 

The retrieval process searches the memory (in models, the base of analogical episodes) to find the closest analog(s) for the given target (input, query) analog. The similarity between episodes is used as a criterion for the search. 
Once the base analogical episode(s) closest to the target is identified, the mapping process finds the correspondences between the elements of two analogical episodes: the target and the base ones. 
The inference process concerns the transfer of the knowledge from the base analogical episode(s) to the target analogical episode. This new knowledge obtained about the target may, e.g., provide the solution to the problem specified by the target analogical episode. 
Because the analogical reasoning is not a deductive mechanism, the candidate inferences are only hypotheses and must be evaluated and checked (see, e.g.,~\cite{GentnerAnalogical2010} and references therein).

Cognitive science has identified quite a few properties demonstrated by subjects when performing analogical reasoning. 
Two types of similarity that influence processing of analogical episodes are distinguished.
Structural similarity (which should not be confused with the structural similarity in HDC/VSA) reflects how the elements of analogs are arranged with respect to each other, that is, in terms of the relations between the elements~\cite{EliasmithAnalogical2001, GentnerAnalogical2012, GentnerAnalogical2017}.
Analogs are also matched by the ``surface'' or ``superficial'' similarity~\cite{GentnerStructure1983, ForbusMAC1995} based on common analogs' elements or a broader ``semantic'' similarity~\cite{HummelDistributed1997, ThagardAnalog1990, EliasmithAnalogical2001}, based on, e.g., joint membership in a taxonomic category
or on similarity of characteristic feature vectors. 
Experiments based on human assessment of similarities and analogies confirmed that both surface (semantic) and structural similarity are necessary for sound retrieval~\cite{ForbusMAC1995}. The structural similarity in the retrieval process is considered less important than that in the mapping process, however, the models of retrieval that take into account only the surface similarity are considered inadequate. 
These properties are expected to be demonstrated by the computational models of analogy~\cite{GentnerComputational2011}. 
In the following sections we discuss applications of HDC/VSA for analogical reasoning.

\paragraph{Analogical retrieval}
\label{sec:analogical:retrieval}

It is known that humans retrieve some types of analogs more readily than others. Psychologists identified different types of similarities and ordered them according to the ease of retrieval~\cite{GentnerStructure1983, RossSimilarities1989, WhartonSimilarity1994, ForbusMAC1995}. The similarity types are summarized in Table~\ref{tab:analog:epissodes}, relative to the base analogical episode. Simple examples of animal stories (adapted by Plate from~\cite{ThagardAnalog1990}) with those similarity types are also presented. All analogical episodes have the same first-order relations (in the example, \texttt{bite()} and \texttt{flee()}). 
There are also higher-order relations \texttt{cause(bite, flee)} and \texttt{cause(flee, bite)} and attribute relations (\texttt{dog, human, mouse, cat}). 

In addition to common first-order relations, the literal similarity (LS) also assumes both the same higher-order relations (in the example, the single relation \texttt{cause(bite, flee)}) and object attributes. The true analogy (AN) has the same higher-order relations, but different object attributes. The surface features (SF) has the same object attributes, but different higher-order relations. The first-order only relations (FOR) differ in both higher-order relations and attributes.
For the analogical retrieval, it is believed that the retrievability order is expressed as LS $\geq$ SF $>$ AN $\geq$ FOR~\cite{GentnerStructure1983, RossSimilarities1989, WhartonSimilarity1994, ForbusMAC1995}.

\begin{table}[tb]
\tiny
\renewcommand{\arraystretch}{1.0}
\caption{Types of analogical similarity.}
\label{tab:analog:epissodes}
    \begin{center}
    \begin{tabular}{c c c c c } 
     \makecell{Similarity type }  & \makecell{Common \\ 1st-order \\ relations } & \makecell{Common \\ high-order \\ relations } & \makecell{Common \\ object \\ attributes } & \makecell{ Examples using \\ animal episodes }  \\ \hline 
    Base &  &   &   &  \makecell{ $\texttt{dog(Spot)}$; $\texttt{human(Jane)}$; \\ $\texttt{cause(bite(Spot, Jane)}$, $\texttt{flee(Jane, Spot))}$ } \\ \hline 
    Literal Similarity (LS) & {\color{cadmiumgreen} \cmark} &  {\color{cadmiumgreen} \cmark} & {\color{cadmiumgreen} \cmark}  &  \makecell{ \texttt{dog(Fido)}; \texttt{human(John)}; \\ \texttt{cause(bite(Fido, John)}, \texttt{flee(John, Fido))} } \\ \hline      
    Surface Features (SF) & {\color{cadmiumgreen} \cmark} &  {\color{red} \xmark} & {\color{cadmiumgreen} \cmark}  &  \makecell{ \texttt{dog(Fido)}; \texttt{human(John)}; \\ \texttt{cause(flee(John, Fido)}, \texttt{bite(Fido, John))} } \\ \hline    
    True Analogy (AN) & {\color{cadmiumgreen} \cmark} &  {\color{cadmiumgreen} \cmark} & {\color{red} \xmark}  &  \makecell{ \texttt{mouse(Mort)}; \texttt{cat(Felix)}; \\ \texttt{cause(bite(Felix, Mort)}, \texttt{flee(Mort, Felix))}  } \\ \hline 
    First-order Only Relations (FOR) & {\color{cadmiumgreen} \cmark} &  {\color{red} \xmark} & {\color{red} \xmark}  &  \makecell{ \texttt{mouse(Mort)}; \texttt{cat(Felix)}; \\ \texttt{cause(flee(Mort, Felix)}, \texttt{bite(Felix, Mort))} } \\ \hline 
     \end{tabular}
    \end{center}
\end{table}

Researchers studying analogical reasoning proposed a number of heuristics-based models of the analogical retrieval. The most influential of them are still MAC/FAC (many are called but few are chosen), which operate with symbolic structures~\cite{ForbusMAC1995}, and ARCS (analog retrieval by constraint satisfaction) using localist neural network structures~\cite{ThagardAnalog1990}. The structure of analogical episodes should be taken into account in their similarity estimation. This requires alignment and finding correspondences between elements of the analogical episodes, as in the mapping (Section~\ref{sec:anlogical:mapping} below), which is computationally expensive. Moreover, unlike for mapping, where only two analogical episodes are considered, in the retrieval process alignment should be repeated for the target analogical episode and each of the base analogical episodes, making such an implementation of the retrieval prohibitive. 

To reduce the computational costs, a two-stage filter and refine (F\&R) approach is used in the traditional models of analogical retrieval.
At the filtering step, the target  analogical episode is compared to all the base analogical episodes using a low-cost similarity of their feature vectors (that only counts the frequency of symbolic names in the analogical episodes, without taking into account the structure). The most similar base analogical episodes are selected as prospective candidates. At the refining step, the candidates are then compared to the target analogical episode by the value of their structural similarity. 
Computationally expensive mapping algorithms (Section~\ref{sec:anlogical:mapping} below) are used for calculating the structural similarity. As the final result, the analogical episodes with the highest structural similarity are returned.

HDC/VSA have been applied to modeling of the analogical retrieval by Plate (see, e.g.,~\cite{PlateAnalogical1994, PlateNested1994, PlateAnalogy2000, PlateHolographic2003}  and references therein). In HDC/VSA, both the set of structure elements and their arrangements influence their HV similarity, so that similar structures (in this case, analogical episodes) produce similar HVs. Because the HV similarity measures are not computationally expensive, the two-stage F\&R approach of the traditional models is not needed. Using HRR, it was shown that the results obtained by a single-stage HV similarity estimation are consistent with both the empirical results in psychological experiments as well as the aforementioned leading traditional models of the analogical retrieval. Note that Plate experimented with analogical episodes different from those tested in the leading models, but they still belong to the proper similarity types, as shown in Table~\ref{tab:analog:epissodes}. Similar results were also reported in~\cite{RachkovskijStructures2001} for SBDR using Plate's episodes.

The study \cite{RachkovskijAnalogy2012} applied SBDR  to represent the analogical episodes in the manner of  \cite{RachkovskijStructures2001}. However, the performance was evaluated using the test bases of the most advanced models of analogical retrieval. The results demonstrated some increase in the recall and a noticeable increase in the precision compared to the leading traditional (two-stage) models. The authors also compared the computational complexity and found that in most of the cases the HDC/VSA approach had advantages over the traditional models.

\paragraph{Analogical mapping}
\label{sec:anlogical:mapping}

The most influential models of analogical mapping include the Structure Mapping Engine (SME)~\cite{FalkenhainerEngine1989} and its versions and further developments~\cite{ForbusExtendingSME2017}, as well as the Analogical Constraint Mapping  Engine (ACME)~\cite{HolyoakAnalogical1989}. SME is a symbolic model that uses a local-to-global alignment process to determine correspondences between the elements of analogical episodes. SME's drawback is a rather poor account of semantic similarity. Also, structure matching in SME is computationally expensive, 
so it is prohibitively expensive to use SME during retrieval
by comparing input to each of (many) analogical episodes in the memory containing all analogical episodes for a structure-sensitive comparison.

ACME is a localist connectionist model that determines analogical mappings using a parallel constraint satisfaction network. Unlike SME, ACME relies not only on the structural information, but also takes into account semantic and pragmatic constraints. ACME is usually even more computationally expensive than SME. 

Further models of mapping based on HRR and BSC were proposed that use techniques based on holistic transformations (Section~\ref{sec:transformation}). One of the limitations of these studies is that the approach was not demonstrated to be scalable to large analogical episodes. 
HRR was also used in another model for the analogical mapping \cite{EliasmithAnalogical2001} called DRAMA, where the similarity between HVs was used to initialize a localist network involved in the mapping.

In~\cite{RachkovskijStructures2001}, similarity of HVs (formed with SBDR) of the analogical episodes was used for their mapping. However, this technique worked only for the most straightforward mapping cases. In~\cite{RachkovskijAnalogical2004}, several alternative techniques for mapping with SBDR were proposed (including direct similarity mapping, re-representation by substitution of identical HVs, and parallel traversing of structures using higher-level roles) and some of them were demonstrated on complex analogies. However, the techniques are rather sophisticated and used sequential operations. 

In \cite{SlipchenkoAnalogical2009}, a kind of re-representation of an analog's element HVs was proposed to allow the analogical mapping  of the resultant HVs on the basis of similarity. The re-representation approach included the superposition of two HVs. One of those HVs was obtained as the HV for the episode's element using the usual representational scheme of the episode, e.g., the HV used for the retrieval (compositional structure representation by HVs, see Section~\ref{PartI-sec:compositional} in~\cite{KleykoSurveyVSA2021Part1}). This HV took into account the semantic similarity. The other HV was the superposition of HVs of elements higher-level roles. This took into account the structural similarity. The proposed procedure was tested in several experiments on simple analogical episodes used in previous studies (e.g., in \cite{PlateHolographic2003, RachkovskijAnalogical2004}, and on rather complex analogical episodes previously used only in the leading state-of-the-art models, e.g., ``Water Flow-Heat Flow'', ``Solar System-Atom'', and ``Schools''~\cite{GentnerStructure1983,HolyoakAnalogical1989}.
It produced the correct mapping results. The analogical inference was also considered. The computational complexity of the proposed approach was rather low and was largely affected by the dimensionality of HVs. 

The problem with the current models of HDC/VSA for analogical mapping is that they lack interaction and competition of consistent alternative mappings. 
They could probably be improved by using an approach involving the associative memory akin to~\cite{GaylerIsomorphism2009}.

Finally, an important aspect of HDC/VSA usage for analogical mapping and reasoning is the compatibility with the existing well-established formats of knowledge representation. This will facilitate the unification of symbolic and subsymbolic approaches for cognitive modeling and Artificial Intelligence.
The work in~\cite{MercierOntology2021} presented a proof of concept of the mapping between Resource Description Framework Schema ontology and HDC/VSA.

\paragraph{Graph isomorphism}

The ability to identify some form of ``structural isomorphism''  is an important component of analogical mapping~\cite{FalkenhainerEngine1989,PlateHolographic2003}. The abstract formulation of isomorphism is graph isomorphism. In~\cite{GaylerIsomorphism2009}, an interesting scheme was proposed for finding the graph isomorphism with HDC/VSA and associative memory. The scheme used the mechanism proposed in \cite{LevyLateralInhibition2009}. The paper presented the HDC/VSA-based implementation of the algorithm proposed in~\cite{PelilloReplicator1999}, which used replicator equations and treated the problem as a maximal-clique-finding problem. In essence, the HDC/VSA-based implementation transformed the continuous optimization problem into a high-dimensional space where all aspects of the problem to be solved were represented as HVs. 
The simulation study (unfortunately performed on a simple graph) showed that the distributed version of the algorithm using the mechanism from \cite{LevyLateralInhibition2009} mimics the dynamics of the localist implementation from~\cite{PelilloReplicator1999}.

\subsubsection{Cognitive modeling}
\label{sec:cog:modeling}

In this part, we briefly cover known examples of using HDC/VSA for modeling particular cognitive capabilities, such as sequence memorization or problem-solving, for cognitive tasks like the Wason task~\cite{EliasmithCognition2005}, n-back task~\cite{GosmannNback2015}, Wisconsin card sorting test~\cite{KajicWisconsin2021}, Raven's Progressive Matrices~\cite{RasmussenInductiveReasoning2011}, or the Tower of Hanoi~\cite{StewartHanoi2011}.

\paragraph{HDC/VSA as a component of cognitive models}

As argued in~\cite{KellyCognition2012}, HDC/VSA is an important tool for cognitive modeling.   
In cognitive science, HDC/VSA has been commonly used as a part of computational models replicating experimental data obtained from humans. For example, in \cite{BlouwConcepts2016} HVs were used as the representational scheme of a computational model. The model was tested using categorization studies considering three competing accounts of concepts: ``prototype theory'', ``exemplar theory'', and ``theory theory''.
The model was shown to be able to replicate experimental data from categorization studies for each of the accounts.
It is also worth mentioning that there are numerous works using context HVs (Section~\ref{sec:context:HVs}) to form models replicating the results obtained in language-related cognitive studies, see, e.g.,~\cite{JonesMeaning2007, ChubalaRecoding2013,JohnsStructure2015, RecchiaSemantic2015, JamiesonITS2018, JohnsCrowdsourced2019, johns2019influence, TalerFluency2020, SchubertLanguage2020}.

\paragraph{Modeling human working memory with HDC/VSA}

A topic that was studied by different research groups working on HDC/VSA is sequence memorization and recall. For example, it was demonstrated in~\cite{ HannaganHolographic2011} (see Section~\ref{PartI-sec:sequences} in~\cite{KleykoSurveyVSA2021Part1}) that a HDC/VSA-based representation of sequences performed better than localist representations when compared on the standard benchmark for behavioral effects. Some studies~\cite{murdock_theory_1982, metcalfe_eich_composite_1982, ChooSerialOrderRecall2010, BlouwWordOrder2013, FranklinMemory2015, KellyDeclarativeMemory2020, GosmannUnifiedSpikingModel2020, ReimannModellingSerial2021} demonstrated how the recall of sequences represented in HVs (Section~\ref{PartI-sec:sequences} in~\cite{KleykoSurveyVSA2021Part1}), albeit with slightly different encoding, can reproduce the performance of human subjects on remembering sequences. This is profound as it demonstrates that simple HDC/VSA operations can reproduce basic experimental findings of human memory studies. An alternative model was proposed in~\cite{CalmusBindingNeurobiologically2019}. Importantly, this work linked the neuroscience literature to the modeling of memorizing sequences with HDC/VSA.

\paragraph{Raven's Progressive Matrices}

Raven's Progressive Matrices is a nonverbal test used to measure general human intelligence and abstract reasoning. Some simple geometrical figures are placed in a matrix of panels, and the task is to select the figure for an empty panel from a set of possibilities~\cite{LovettRaven2010}. 
Note that the key work related to Raven's progressive matrices goes back to the 1930s (see~\cite{Raven2000}). It was widely used to test human fluid intelligence and abstract reasoning since the 1990s~\cite{Carpenter1990}. 
The task was first brought to the scope of HDC/VSA in~\cite{RasmussenInductiveReasoning2011, RasmussenSpiking2014} using HRR and its subsequent implementation in spiking neurons. Later studies~\cite{EmruliAnalogical2013,LevyBeetle2014} demonstrated that other HDC/VSA models (BSC and MAP, respectively) can also be used to create systems capable of solving a limited set of Raven's Progressive Matrices containing only the progression rule. 

In all studies the key ingredient of the solution was the representation of a geometrical panel by its HV (e.g., giving access to the symbolic representation of the panel followed by using role-filler bindings of the shapes and their quantity present in the panel). Subsequently, the HV corresponding to the transformation between the HVs of the adjacent pairs of panels was obtained using the ideas from~\cite{NeumannTransformation2002} (see Section~\ref{sec:transformation}). The transformation HV was then used to form a prediction HV for the blank panel in the matrix. The candidate answer with the HV most similar to the prediction HV was then chosen as the answer to the test. 
All of the previously described studies have two limitations: first, they assume the perception system provides the symbolic representations that support the reasoning for solving Raven's Progressive Matrices test; and second, they only support the progression rule. 
A recent work~\cite{herscheNVSA2022} addressed these limitations by positioning VSA/HDC as a common language between a neural network (to solve the perception issue) and a symbolic logical reasoning engine (to support more rules). 
Specifically, it exploited the superposition of multiplicative bindings in a neural network to describe raw sensory visual objects in a panel and used Fourier Holographic Reduced Representations (FHRR) to efficiently emulate a symbolic logical reasoning with a rich set of rules~\cite{herscheNVSA2022}.

\paragraph{The Tower of Hanoi task}

The Tower of Hanoi task, which is a simple mathematical puzzle, is another example of a task used to assess problem solving capabilities. The task involves three pegs and a fixed number of disks of different sizes with holes in the middle such that they can be placed on the pegs. Given a starting position, the goal is to move the disks to a target position. Only one disk can be moved at a time and a larger disk cannot be placed on top of a smaller disk. 

In~\cite{StewartHanoi2011}, an HDC/VSA-based model capable of solving the Tower of Hanoi tasks was presented. The binding and superposition operations were used to form HVs of the current position and a set of goals identified by the model. The model implemented a known algorithm for solving the task given valid starting and target positions. The performance of the model was compared to that of humans regarding of time delays, which were found to be qualitatively similar.

\paragraph{Modeling the visual similarity of words}

Modeling human perception of word similarity with HDC/VSA was based on the experimental data obtained for human subjects in~\cite{DavisSpatial2010}. 
The task was to model the human patterns of delays in priming tasks with the similarity values of sequence HVs obtained from various HDC/VSA models and for various schemes for sequence representation  (Section~\ref{PartI-sec:sequences} in~\cite{KleykoSurveyVSA2021Part1}). 

In the task of modeling restrictions on the perception of word similarity, four types of restrictions (a total of 20 similarity patterns) were summarized in~\cite{HannaganHolographic2011}.
In~\cite{HannaganHolographic2011}, the BSC model was employed to  represent symbols in their positions; various string representations with substrings were also used. 
Symbols in positions with correlated HVs to represent nearby positions were studied in ~\cite{CohenOrthogonality2013} for the BSC, FHRR and HRR models. 
Their results demonstrated partial satisfaction of the restrictions.  
On the other hand, substring representations from~\cite{CoxHolographic2011} with HRR, and symbols-at-correlated positions representations 
from~\cite{RachkovskijEquivariant2021} with SBDR using permutations as well as the ones from~\cite{RachkovskijRecursiveBinding2022} with FHRR met all the restrictions for certain choices of the similarity measure and values of a scheme's hyperparameters.  
In the task of finding correlations between human and model similarity data,~\cite{RachkovskijEquivariant2021,RachkovskijRecursiveBinding2022} demonstrated results were on a par with those of the string kernel similarity measures from~\cite{HannaganProtein2012}.

\paragraph{General-purpose rule-based and logic-based inference with HVs}

The study in~\cite{StewartSymbolicReasoning2010} presented a spiking neurons model that was positioned as a general-purpose neural controller.
The controller was playing a role analogous to a production system capable of applying inference rules. HDC/VSA and their operations played a key role in the model providing the basis for representing symbols and their relations. The model was demonstrated to perform several tasks: repeating the alphabet, repeating the alphabet starting from a particular letter, and answering simple questions similar to the ones in Section~\ref{sec:holistic:proc}.
Another realization of a production system with the Tensor Product Representations model was demonstrated in~\cite{SmolenskyProduction1989}.
Several examples of hierarchical reasoning using the superposition operation on context HVs representing hyponyms to form representations of hypernyms were presented in~\cite{KommersHierarchicalReasoning2015}.

In~\cite{SummersInference2020}, it was demonstrated how HVs can be used to represent a knowledge base with clauses for further performing deductive inference on them. The work widely used negation for logical inference, which was also discussed in~\cite{KussulAssociative1992}.
Reasoning with HRR using modus ponens and modus tollens rules was  demonstrated in~\cite{KvasnivckaDeductiveHRR2006}. 
The works in~\cite{SchmidtkeMultimodalActivation2021, SchmidtkeContextLogicVSA2021, SchmidtkeAnalogicalSemantics2022} discussed the usage of VSA/HDC for the realization of context logic language and demonstrated the inference procedures.

\subsubsection{Computer vision and scene analysis}
\label{sec:comp:vision}

This section summarizes different aspects of using HDC/VSA for processing visual data. This is one of the newest and least explored application areas of HDC/VSA.

\paragraph{Visual analogies}

In~\cite{YilmazAnalogy2015c}, a simple analogy-making scenario was demonstrated on 2D images of natural objects (e.g., bird, horse, automobile, etc.). This work took an image representing a particular category, e.g., a bird. 
The HVs of images were obtained through using convolutional neural networks (Section~\ref{PartI-sec:2Dimages:neural:nets} in~\cite{KleykoSurveyVSA2021Part1}) and cellular automata computations (see~\cite{YilmazSymbolic2015} for the method description). Several (e.g., $50$ in \cite{YilmazAnalogy2015c}) such binary HVs (e.g., for images of different birds) were superimposed together to form the HV of a category, e.g.,
\noindent
\begin{equation}
\mathbf{land}=\mathbf{animal} \circ \mathbf{horse} + \mathbf{vehicle} \circ \mathbf{automobile};
\end{equation}
\noindent
\begin{equation}
\mathbf{air}=\mathbf{animal} \circ \mathbf{bird} + \mathbf{vehicle} \circ \mathbf{airplane}.
\end{equation}
\noindent
The category HVs were used to form some statements using the HDC/VSA operations. 
Inspired by the well-known example of ``Dollar of Mexico?'' (as in the techniques of Section~\ref{sec:holistic:proc}) it was shown that one could perform queries of a similar form as ``What is the Automobile of Air?'' ($\mathbf{AoA}$), but using HVs formed from the 2D images:
\noindent
\begin{equation}
\mathbf{AoA}= \mathbf{air} \oslash  (\mathbf{land} \oslash \mathbf{automobile}) .
\end{equation}
\noindent
The system demonstrated a high accuracy (98\%) of correct analogy-making on previously unseen images of automobiles.

\paragraph{Reasoning on visual scenes and Visual Question Answering}

Visual Question Answering is defined as a task where an agent should answer a question about a given visual scene. In \cite{MontoneVQANIPS}, a trainable model was presented that used HDC/VSA for this task. The model in \cite{MontoneVQANIPS} differs from the state-of-the-art solutions that usually include a combination of a recurrent neural network (handles questions and provides answers) and a convolutional neural network (handles visual scenes). 
The model included two parts. The first part transformed a visual scene into an HV describing the scene using a neural network. This part used only one feed-forward neural network, which took a visual scene and returned its HV. The second part of the model defined the item memory of atomic HVs as well as HVs of questions along with their evaluation conditions in terms of cosine similarity thresholds. 
	The neural network was trained to produce HVs associated with a dataset of simple visual scenes (two figures in various combinations of four possible shapes, colors, and positions). The gradient descent used errors from the question answering on the training data, which included five predefined questions. It was shown that the trained network successfully produced HVs that answered questions for new unseen visual scenes. The five considered questions were answered with 100\% accuracy. On previously unseen questions the model demonstrated an accuracy in the range of 60-72\%.

Similar to~\cite{MontoneVQANIPS}, there were attempts in~\cite{MaudgalyaVisualVSA2020, KentVSAScene2017, herscheNVSA2022} to train neural networks to output HVs representing a structured description of scenes (see also Section~\ref{PartI-sec:2Dimages:neural:nets} in~\cite{KleykoSurveyVSA2021Part1}), which could then be used for computing visual analogies.

Another approach to Visual Question Answering  with HDC/VSA was outlined in~\cite{KovalevSemioticHD2020, KovalevVectorSemiotic2021}, where a visual scene was first preprocessed to identify objects and construct a scene data structure called the causal matrix (it stored some object attributes including positions). This data structure describing a scene was transformed into an HV that could then be queried using HDC/VSA operations similar to those from~\cite{MontoneVQANIPS}.  
In~\cite{KirilenkoVQA2021}, it was applied to a dataset constructed to facilitate visual navigation in human-centered environments.
This approach was further extended from Visual Question Answering to Visual Dialog in~\cite{KovalevVisualDialog2021}.

Another application of HDC/VSA for representation and reasoning on visual scenes, similar in its spirit to the Visual Question Answering, was presented in \cite{WeissOlshausenSpatial16}. The approach represented visual scenes in the form of HVs using Fourier HRR. The paper transformed continuous positions in an image to complex-valued HVs such that the spatial relation between positions was preserved in the HVs (see the ``fractional power encoding'' in  Sections~\ref{PartI-sec:scalars:vectors:compos} and~\ref{PartI-sec:2Dimages:role-filler} in~\cite{KleykoSurveyVSA2021Part1}). During the evaluation,  handwritten digits and their positions were identified using a neural network with an attention mechanism. Then the identified information was used to create a complex-valued compositional HV describing the scene. Such scene HVs were used to answer relational queries like ``which digit is below 2 and to the left of 1?''. 

A further exploration of this task was presented in~\cite{LuFractional2019}. The approach in~\cite{WeissOlshausenSpatial16}  also demonstrated solving a simple navigation problem in a maze.    In~\cite{KomerNavigation2020}, navigation tasks in 2D environments were further studied by using HVs as an input to a neural network producing outputs directing the movement. Neural networks trained with HVs demonstrated the best performance amongst methods considered. There was also a recent attempt to implement this continuous representation with spiking neurons~\cite{DumontGrid2020}.

\subsection{Cognitive  architectures}
\label{sec:cog:arch}

HDC/VSA have been used as an important component of several bio-inspired cognitive architectures. Here, we briefly describe these  proposals.

\subsubsection{Semantic Pointer Architecture Unified Network}
\label{sec:cog:spaun}

The most well-known example of a cognitive architecture using HDC/VSA is called ``Spaun'' for Semantic Pointer Architecture Unified Network (see its overview in \cite{EliasmithSPAUN2012} and a detailed description in~\cite{EliasmithBuildBrain2013}). 
Spaun is a large spiking neural network ($2.5$ million neurons). 
Spaun that uses HRR for data representation and manipulation. 
In the architecture, HVs play the role of ``semantic pointers''~\cite{BlouwConcepts2016} aiming to integrate connectionist and symbolic approaches.
It has an ``arm'' for drawing its outputs as well as ``eyes'' for sensing inputs in the form of 2D images (handwritten or typed characters). 
Spaun (without modifying the architecture) was demonstrated in eight different cognitive tasks that require different behaviors. The tasks
used to demonstrate the capabilities of Spaun were: 
copy drawing of handwritten digits, handwritten digit recognition, reinforcement learning on a three-armed bandit task, serial working memory, counting of digits, question answering given a list of handwritten digits, rapid variable creation, and fluid syntactic or semantic reasoning.
The same principles were used in~\cite{CrawfordKnowledge2016} to represent the WordNet knowledge base with HVs allowing enriching, e.g., Spaun, with a memory storing some prior knowledge.

\subsubsection{Associative-Projective Neural Networks}
\label{sec:cognitive:APNN}

The cognitive architecture of Associative-Projective Neural Networks (APNNs) that use the SBDR model was proposed and presented in~\cite{KussulAPNN1991, KussulSequences1991, KussulAssociative1992, KussulNNMM2010, RachkovskijWorldModel2013}.
The goal was to show how to construct a complex hierarchical model of the world, that presumably exists in human's and higher animals' brains, as a step towards Artificial Intelligence. 

Two hierarchy types were considered: the compositional (part-whole) as well as the categorization or generalization (class-instance or is-a) ones. An example of the compositional hierarchy is: letters $\rightarrow$ words $\rightarrow$ sentences $\rightarrow$ paragraphs $\rightarrow$ text. Another example is the description of knowledge base episodes in terms of their elements of various complexity, from attributes to objects to relations to higher-order relations (see Section~\ref{sec:analogical:reasoning}). An example of the categorization hierarchy is: 
dog $\rightarrow$ spaniel $\rightarrow$ spaniel Rover or 
apple $\rightarrow$ big red apple $\rightarrow$ this big red apple in hand. 

The proposed world model relies on models of various modalities, including sensory ones (visual, acoustic, tactile, motoric, etc.) and more abstract modalities (linguistics, planning, reasoning, abstract thinking, etc.) that are organized in hierarchies. The models are required for objects of different nature, e.g., events, real objects, feelings, attributes, etc. Models (their representations) of various modalities can be combined, resulting in multi-modal representations of objects and associating them with the behavioral schemes (reactions to objects or situations), see details in~\cite{KussulAPNN1991, KussulSequences1991, KussulAssociative1992, KussulNNMM2010, RachkovskijWorldModel2013}.

The APNN architecture is based on HDC/VSA (though it was proposed long before the terms HDC and VSA appeared), in particular, models are represented by SBDR (Section~\ref{PartI-sec:SBDR} in~\cite{KleykoSurveyVSA2021Part1}). 
An approach to formation, storage, and modification of hierarchical models was proposed. This is facilitated by the capability to represent in HVs of fixed dimensionality (for items of various complexity and generality) various heterogeneous data types, e.g., numeric data, images, words, sequences, structures (Section~\ref{PartI-sec:data:to:HV} in~\cite{KleykoSurveyVSA2021Part1}). As usual for HDC/VSA, the model HVs can be constructed on-the-fly (without learning). APNNs have a multi-module, multi-level, and multi-modal design. A module forms, stores, and processes many HVs representing models of objects of a certain modality and of a certain level of compositional hierarchy. A module's HVs are constructed from HVs obtained from other modules, such as lower-level modules of the same modality, or from modules of other modalities. The lowest level of the compositional hierarchy consists of modules providing a representation grounding (atomic HVs). 

For SBDR, a HV is similar to the HVs of its elements of a lower compositional hierarchy level, as well as to the HVs of the higher level, of which the HV is an element. So, using similarity search (in the item memory of levels) it is possible to recover both the lower-level element HVs and compositional HVs of the higher level.

Each module has a long-term memory where it stores its HVs. A Hopfield-like distributed auto-associative memory~\cite{GritsenkoAMSurvey2017, FrolovWillshaw2002, FrolovTime2006} was suggested as a module memory. It performs the clean-up procedure for noisy or partial HVs by a similarity search. However, its unique property is the formation of the second main hierarchy type, i.e., of generalization (class-instance) hierarchies. It is formed when many similar (correlated) HVs are stored (memorized), based on the idea of Hebb's cell assemblies including cores (subsets of HV 1-components often occurring together, corresponding to, e.g., typical features of categories and object-prototypes) and fringes (features of specific objects), see~\cite{RachkovskijWorldModel2013, GritsenkoAMSurvey2017}.

It is acknowledged that a world model comprising domain(s) specific knowledge as well as information about an agent itself is necessary for any intelligent behavior. Such a model allows comprehension of the world by an intelligent agent and assists it in its interactions with the environment, e.g., through predictions of action outcomes. 

The main problem with the APNN architecture is that not all its aspects have been modeled. For example, there are the following questions, which do not have exact answers: 
\begin{itemize}
    \item how to extract objects and their parts of various hierarchical levels?
    \item how to determine the hierarchical level an object belongs to?
    \item how to work with an object that may belong to different hierarchy levels and modules?
    \item how to represent objects invariant of their transformations? 
\end{itemize}
Also, modeling cores and fringes formation in distributed auto-associative memory is still fragmentary as of now. 
Finally, it worth noting that similar ideas are currently being developed in the context of deep neural networks~\cite{GhaziRecursive2019, DengHierarchical2021}.

\subsubsection{Hierarchical Temporal Memory}
\label{sec:cognitive:HTM}

An interesting connection between HDC/VSA and a well-known architecture called Hierarchical Temporal Memory (HTM)~\cite{HTM2011} was presented in \cite{PadillaVSAHTM2014}.
The work showed how HTM can be trained in its usual sequential manner to support basic HDC/VSA operations: binding and superposition  of sparse HVs, which are natively used by HTM. 
Even though permutations were not discussed, it is likely that they also could be implemented, so that HTM could be seen as another HDC/VSA model, which additionally has a learning engine in its core.

\subsubsection{Learning Intelligent Distribution Agent}
\label{sec:cognitive:LIDA}

A version of the well-known  symbolic cognitive architecture LIDA (Learning Intelligent Distribution Agent ~\cite{FranklinLIDA2013}),  working with HVs, was presented in~\cite{SnaiderVLIDA2014}.
In particular, the Modular Composite Representations model was used~\cite{SnaiderModular2014}.    
Moreover, memory mechanisms in the proposed architecture were also related to HDC/VSA: 
an extension of Sparse Distributed Memory~\cite{KanervaSDM1988}, known as Integer Sparse Distributed Memory~\cite{SnaiderIntegerSDM2013}, was used.
The usage of HVs allowed resolving  some of the issues with the original model~\cite{FranklinLIDA2013}, which relies on directed graph-like structures  such as representation capability, flexibility, and scalability.

\subsubsection{Memories for cognitive  architectures}
\label{sec:cognitive:memory}

Memory is one of the key components of any cognitive architecture and for modeling cognitive abilities of humans. There is a plethora of memory models. For example, MINERVA 2~\cite{HintzmanMinerva1984} is an influential computational model of long-term memory. However, in its original formulation MINERVA 2 was not very suitable for an implementation in connectionist systems. Nevertheless, it was demonstrated in~\cite{KellyTesseract2017} that the Tensor Product Representations model (Section~\ref{PartI-sec:framework:TPR} in~\cite{KleykoSurveyVSA2021Part1}) can be used to formalize MINERVA 2 as a fixed size tensor of order four. Moreover, it was demonstrated that the lateral inhibition mechanism for HDC/VSA~\cite{GaylerIsomorphism2009} and HRR can be used to approximate MINERVA 2 with HVs. HVs allowed to compress the exact formulation of the model, which relies on tensors, into several HVs, thus making the model more computationally tractable at the cost of lossy representation in HVs. 

Another example of using HVs (with HRR) for representing concepts is a Holographic Declarative Memory~\cite{KellyDeclarativeMemory2015, AroraMind2018,KellyDeclarativeMemory2020} related to BEAGLE~\cite{JonesMeaning2007} (see Section~\ref{sec:BEAGLE}). It was proposed as a declarative and procedural memory in cognitive architectures. It was shown that the memory can account for many effects such as primacy, recency, probability estimation, interference between memories, and others.

In~\cite{JohnsGenerating2015, JamiesonITS2018} BEAGLE (Section~\ref{sec:BEAGLE}) was extended to store (instead of one context HV per word) episodic memory of the observed data as HVs of all the contexts. This extension was called the instance theory of semantics. Each word was represented by an atomic random HV. A word's context (a sentence) HV is constructed as a superposition of its word HVs and is stored in the memory. The HV of some query word is constructed as follows. First, the $\text{sim}_{\text{cos}} $ of the query word HV and each context HV is calculated and raised to a power, producing a vector of ``trace activations''. Then, context HVs are weighted by traces and summed to produce the retrieved (``semantic'') HV of the query word.  

The study in~\cite{CrumpRetrieval2020} introduced a ``weighted expectancy subtraction'' mechanism that formed actual context HV as follows. First, the context HV produced the retrieved HV as explained above. Then, the retrieved HV was weighted and subtracted from the initial context HV. During the retrieval, the weighted HV of the second retrieval iteration was subtracted from the HV of the first retrieval iteration. This allowed flexibly controlling the construction of general versus specific semantic knowledge. 
The work in~\cite{OrorbiaKellyCogNGen2022} proposed CogNGen - a core of a cognitive architecture that combines predictive processing based on neural generative coding and HDC/VSA models of human memory. The CogNGen architecture learns across diverse tasks and models human performance at larger scales.

\section{Discussion}
\label{sec:disc}

\subsection{Application areas}
\subsubsection{Context HVs}
When it comes to context HVs, Random Indexing and Bound Encoding of the Aggregate Language Environment have appeared as improvements to Latent Semantic Analysis – e.g., they do not require Singular Value Decomposition and can naturally take order information into account. However, they were largely overshadowed after the introduction of ``neural'' word embeddings in Word2vec~\cite{MikolovWord2vec2013} or GloVe~\cite{pennington2014glove}. The latter are the result of an iterative process, which takes numerous passes via training data to converge.
At the same time, an important fact is that distributional models such as Latent Semantic Analysis can in fact benefit from some techniques used in neural word embeddings \cite{levy2015improving}. 
Concerning, e.g., Bound Encoding of the Aggregate Language Environment, as recently demonstrated in~\cite{JohnsNegative2019}, the method can benefit from negative information. Nevertheless, the current de facto situation in the natural language processing community is that Bound Encoding of the Aggregate Language Environment and Random Indexing methods are rarely the first candidates when it comes to choosing word embeddings. On the other hand, since Bound Encoding of the Aggregate Language Environment has been proposed within cognitive science community, it still plays an important role in modeling cognitive phenomena related to memory and semantics~\cite{JamiesonITS2018, CrumpRetrieval2020}.
Also, in contrast to the iterative methods, Random Indexing and Bound Encoding of the Aggregate Language Environment only require a single pass through the training data to form context HVs. In some situations, this could be an advantage, especially since the natural language processing community is becoming increasingly concerned about the computational costs of algorithms~\cite{StrubellEnergyNLP}.

\subsubsection{Classification}
While right now the classification with HDC/VSA is flourishing, there are still important aspects that are often not taken into account in these studies. 

\paragraph{Formation of HVs}
An important aspect of the formation of HVs is the initial extraction of features from raw data such as 2D images or acoustic signals.
Usually, directly transforming such data into HVs does not result in a good performance, so an additional step of extracting meaningful features is required.

Another important aspect is that, when constructing HVs from feature data for classification, in most cases the transformation of data into HVs is somewhat ad hoc. While there likely will not be a straightforward answer to how transforming data into HVs, it is still important to mention several issues. 

It is a well-known fact that the advantage of nonlinear transformation is that classes not linearly separable in the original representation, might become linearly separable after a proper nonlinear transformation to a high-dimensional space (often called lift). This allows using not only $k$-Nearest Neighbor classifiers, but also well-developed linear classifiers to solve problems that are not linearly separable. So, nonlinearity is an important aspect of transforming data into HVs. All transformations of data into HVs that we are aware of seem to be nonlinear. 
However, there are no studies that scrutinize and characterize the nonlinearity properties of HVs obtained from the compositional approach. Moreover, most of the studies choose a particular transformation of numeric vector and stick to it. 
One of the most common choices is randomized ``float'' coding~\cite{RachkovskijScalars2005, WiddowsContinuous2015, KleykoTradeoffs2018, RahimiBiosignal2019}.
There is, however, a recent study~\cite{FradyFunctions2021,FradyFunctionsNICE2022} that established a promising connection between kernel methods~\cite{ScholkopfKernels2002,rahimi2007random} and the fractional power encoding for representing numeric vectors 
as well as an earlier algorithm for approximating a particular type of kernels (tree kernels) with the HRR model~\cite{ZanzottoTree2012}.
In our opinion, the transformation of data into HVs is a hyperparameter of the model and using, e.g., cross-validation to choose the most promising transformation will likely be the best strategy when considering a range of different datasets.

\paragraph{Choice of classifier}
\label{sec:disc:class:classifier}

As we saw in Section~\ref{sec:app:class}, centroids are probably the most common approach to forming a classifier in HDC/VSA. This is understandable, since centroids have an important advantage in terms of computational costs -- they are very easy to compute. However, as pointed out in~\cite{KarlgrenSemantics2021}, the result of superposition does not provide generalization in itself, it is just a representation of combinations of HVs of training samples.  
Practically, it means that the centroids are not the best performing approach when it comes to classification performance. 
One way to improve the performance is to assign weights when including new samples into centroids~\cite{RahimiBiosignal2016,HernandezOnlineHD2021}. It was also shown that the perceptron learning rule in~\cite{ImaniVoiceHD2017} and loss-based objective in~\cite{HerscheContinualLearn2022} might significantly improve centroids. Earlier work on HV-based classifiers, e.g., \cite{KussulAdaptive1993,KussulThreshold1994, RachkovskijDatagen1998, RachkovskijClassifiers2007} also used linear perceptron and Support Vector Machine classifiers with encouraging results. 
Note that a large-margin perceptron usually trains much faster than a Support Vector Machine for big data, while providing classification quality at the same level as the Support Vector Machine and usually much higher than that of the standard perceptron. 
Another recent result~\cite{DiaoGLVQHD2021} is that centroids can be easily combined with a known conventional classifier --  generalized learning vector quantization~\cite{SatoGLVQ1995}. Using an HDC/VSA transformation of data to HV, the authors obtained state-of-the-art classification results on a benchmark~\cite{HundredsClassifiers2014}. 
In general, we believe that when inventing new mechanisms of classification with HDC/VSA, it is important to report the results on collections of datasets instead of only a handful of datasets. For example, for feature-based classification the UCI Machine Learning Repository~\cite{DuaUCI2019} and subsets thereof (e.g.,~\cite{HundredsClassifiers2014}) are a common choice (examples of HDC/VSA using it are~\cite{KleykoDensityEncoding2020,DiaoGLVQHD2021, FradySDR2020}).  
For univariate temporal signals, the UCR Time Series Archive~\cite{DauUCRSeries2019} is a good option that was used, e.g., in~\cite{SchlegelHDC-MiniROCKET2022}. 
If a reported mechanism targets a more specific application area, it would be desirable to evaluate it on a relevant collection for that area.

Many other types of classifiers such as $k$-Nearest Neighbors~\cite{KarunaratneHDAugmented2021} are also likely to work with HVs as their input. 
When HVs are generated by a nonlinear transformation (in~\cite{KarunaratneHDAugmented2021}, using an HDC/VSA-guided convolutional neural network feature extractor), the $k$-Nearest Neighbors classifier forms an ``explicit memory'' in memory-augmented neural networks or Neural Turing Machines~\cite{NTMGraves2014}. The contents of the explicit memory can be compressed using outer products with randomized labels~\cite{KleykoAugmented2022}. 
As mentioned in the previous section, linear classifiers can be used with nonlinearly transformed HVs.
For example, the ridge regression, which is commonly used for randomized neural networks~\cite{RCNNSsurvey}, performed well with HVs~\cite{RosatoHDDistributed2021}. 
However, not all conventional classifiers work well with HVs~\cite{AlonsoHyperEmbed2020}. That is because some of the algorithms (e.g., decision tree or Naïve Bayes) assume that any component of a vector can be interpreted on its own. It is a reasonable assumption when components of vectors are meaningful features, but in HVs a component does not usually have a meaningful interpretation. 
In the case of HDC/VSA with sparse representations, special attention should be given to classifiers that benefit from sparsity. Examples of such classifiers are the sparse Support Vector Machine~\cite{EshghiSparseSVM2016} and winnow algorithm~\cite{LittlestoneWinnow1988}.

\paragraph{Applications in machine learning beyond classification}

There are also efforts to apply HDC/VSA within machine learning outside of classification.
Examples of such efforts are using data transformed into HVs for clustering~\cite{BandaragodaTrajectoryTraffic2019, HernandezClustering2021}, unsupervised learning\cite{MirusAbnormal2020, OsipovHyperSeed2021}, 
multi-task learning~\cite{ChangHDTaskProjected2020, ChangHDInformationPreserved2020,ChangMulTaHDC2021}, distributed learning~\cite{RosatoHDDistributed2021, HsiehFL2021}, model compression~\cite{HerscheCompressingBCI2020,ChangMulTaHDC2021, RosatoHDDistributed2021,RosatoCompression2021}, and ensemble learning~\cite{EnsembleBurrello2021,EnHDCWang2022}.

It is expected that fractional power encoding~\cite{PlateNested1994,KomerContinuous2019, FradyFunctions2021} (Section~\ref{PartI-sec:scalars:vectors:compos} in~\cite{KleykoSurveyVSA2021Part1}) is going to be a particularly fruitful method for enabling new applications beyond classification.  
This expectation is based on two facts.
First, fractional power encoding is known to approximate kernels, which allows for an efficient implementation of kernel methods. 
There are already examples of its use to implement methods for probability density estimation~\cite{FradyFunctions2021, FradyFunctionsNICE2022}, kernel regression~\cite{FradyFunctions2021, FradyFunctionsNICE2022}, Gaussian processes-based mutual information exploration~\cite{FurlongSSPMI2022}, representing probability  statements~\cite{FurlongProbability2022}, path integration~\cite{DumontPath2022}, and reinforcement learning~\cite{BartlettReinforcement2022}.
Second, fractional power encoding provides a simple but powerful way for representing numeric data in HVs, which allows numerous applications relying on such data. 
Some recent examples include simulation and prediction of dynamical systems~\cite{VoelkerFPEDynamical2021}, reasoning on 2D images~\cite{WeissOlshausenSpatial16, FradyDisentangling2018, LuFractional2019}, 
navigation in 2D environments~\cite{WeissOlshausenSpatial16,KomerNavigation2020}, representation of time series~\cite{SchlegelHDC-MiniROCKET2022}, 
and even modeling in neuroscience~\cite{FradyFramework2018, DumontGrid2020}.

We did not devote separate sections to these efforts as the studies are still scarce, but the interested readers are kindly referred to the above works for the initial investigations on these topics.

\subsubsection{Real-world use-cases and new application areas}

Section~\ref{sec:applications} demonstrated that there have been numerous attempts to apply HDC/VSA in a diverse range of scenarios spanning from communications to analogical reasoning. As we can see from Section~\ref{sec:app:class}, the most recent uptick in the research activity was applying HDC/VSA to classification tasks. In the near future, we are likely to see them being applied to solving classification tasks in new domains. 
Examples of such new domains recently appeared in, e.g.,~\cite{VougioukasBranch2021}, where HDC/VSA was applied to branch prediction in processor cores and~\cite{ShahroodiDemeter2022}, where HDC/VSA was applied to food profiling.

Concerning the applications of analogical reasoning (Section~\ref{sec:analogical:reasoning}), the major bottleneck is still the transformation of textual, speech or pictorial descriptions of analogical episodes to directed ordered acyclic graphs that can then be transformed into HVs (Section~\ref{PartI-sect:graphs:labelled} in~\cite{KleykoSurveyVSA2021Part1}). Note that this problem concerns not only analogical reasoning based on HDC/VSA, but all methods that use predicate-based descriptions as inputs. 

Nevertheless, there is still a considerable way to go in order to demonstrate how HDC/VSA-based solutions scale up to real-world problems. We, however, strongly  believe that, similarly to the modern reincarnation of connectionist models, eventually research will distill the niches where the advantages of HDC/VSA are self-evident. Currently, one promising niche seems to be the time series classification~\cite{SchlegelHDC-MiniROCKET2022}, particularly in-sensor classification of biomedical signals~\cite{MoinWearable2021} and prosthetic grasping~\cite{OlascoagaProsthetic2022}.
Furthermore, the exploration of HDC/VSA in novel application domains should be continued. For instance, there were recent applications in communication~\cite{KimHDM2018,HsuCollisionTolerant2019, HsuNonOrthogonalModulation2020,HerscheHDMClassifier2021} and in distributed systems~\cite{SimpkinHDWorkflow2019} (see Section~\ref{sec:stor:trans:comm}), which were not foreseen by the community. 
Another recent example is the attempt to apply HDC/VSA to robotics problems~\cite{MendesRobotSDM2008, MendesRobotSDM2012, TravnikRepresenting2017, NeubertRobotics2019, MitrokhinSensorimotor2019, McDonaldHDRobotics2021, NeubertAggregation2021}.

\subsection{Interplay with neural networks}

\subsubsection{HVs as input to neural networks}

One of the most obvious ways to make an interplay between HDC/VSA and neural networks is by using HVs to represent input to neural networks. This is a rather natural combination of the two, because, in essence, neural networks often work with distributed representations. So, processing information distributed in HVs is not an issue for neural networks. However, since HVs are high-dimensional, it is not always possible to use them as the input: the size of neural networks' input layer should be set to $D$ (e.g., in~\cite{MaHolisticMemoriz2018}, whereby a fully-connected ``readout''  layer for a task of choice was trained on $D$-dimensional input HVs)  or even to a tensor comprised of HVs (e.g., to represent each position in the retina by its HV, without superposition of HVs).
Moreover, the local structure of the input signal space may become different from that used, e.g., in convolutional neural networks. This could require very different neural network architectures compared to modern deep neural networks. 

There are, nevertheless, scenarios where using HVs with neural networks appeared to be beneficial.
First, HVs are useful in situations when the data to be fed to a neural network are high-dimensional and sparse. Then HVs can be used to form more compact distributed representations of these data. A typical example of using such high-dimensional and sparse data is $n$-gram statistics. There are works which studied tradeoffs between the dimensionality of HVs representing $n$-gram statistics (see Section~\ref{PartI-sec:sequences:ngrams} in~\cite{KleykoSurveyVSA2021Part1}) and the performance of neural networks using these HVs as their input~\cite{KleykoBoostingSOM2019, AlonsoHyperEmbed2020}. 
These works demonstrated that it is possible to achieve the same or very similar classification performance with networks of much smaller size. 
Moreover, the degradation of the classification performance is gradual with the decreasing size of HVs, so their dimensionality can be used to control the tradeoff between the size of the network and its performance. 
On top of creating more compact representations, an additional advantage of HVs might lie in making HVs binary as in, e.g., Binary Spatter Codes. This might be leveraged in situations where the whole model is binarized~\cite{ShridharEnd2End2020}.

Also, HVs may be useful when the size of input is not fixed but instead could vary for different inputs. Since neural networks are not flexible in changing their architecture, HDC/VSA can be used to take care of forming fixed size HVs for input of variable size.
This mode of a neural network interface has been demonstrated in an automotive context to represent either varying number of intersections being crossed by a vehicle~\cite{BandaragodaTrajectoryTraffic2019} or the dynamically changing environment around a vehicle~\cite{MirusBalanced2020, MirusBehavior2019}. 
Further promising avenues for this mode are graphs and natural language processing, since there is a lot of structure in both, which can potentially be represented in HVs~\cite{MaHolisticMemoriz2018,KarlgrenUtterances2019}. Some investigations in this direction using Tensor Product Representations were presented in~\cite{ChenTPRMapping2020}.

We foresee that this interface mode might expand the applicability of neural networks, as it allows relieving the pressure of forming the task either with fixed size input or in the form of, e.g., a sequence suitable for recurrent neural networks. However, it may require a replacement of the widely used convolutional layers. Although, there are new results~\cite{TolstikhinMLP2021} suggesting that fully connected neural networks might be a good architecture even for vision tasks.

\subsubsection{The use of neural networks for producing HVs}
\label{disc:ANNs:produce}

Transforming data into HVs (see Section~\ref{PartI-sec:data:to:HV} in~\cite{KleykoSurveyVSA2021Part1}) might be a non-trivial task, especially when data are unstructured and of non-symbolic nature as, e.g., in the case of images (Sections~\ref{PartI-sec:2D:permute} and~\ref{PartI-sec:2Dimages:role-filler} in~\cite{KleykoSurveyVSA2021Part1}). Also, those transformations are usually not learned. This challenge stimulates the interface between neural networks and HDC/VSA in the other direction, i.e., to transform activations of neural network layer(s) into HVs. For example, as mentioned in Section~\ref{PartI-sec:2Dimages:neural:nets} in~\cite{KleykoSurveyVSA2021Part1}, it is very common to use activations of convolutional neural networks to form HVs of images. This is commonly done using standard pre-trained neural networks~\cite{YilmazMachine2015, MitrokhinCNN2020, NeubertAggregation2021}. Two challenges here are to increase the dimensionality and change the format of the neural network representations to conform with the HV format requirements.  
The former one is generally addressed by expanding the dimensionality, e.g., by random projection, possibly with a subsequent binarization by thresholding~\cite{HerscheBinarization2020, NeubertAggregation2021}.
Some neural networks already produce binary vectors (see~\cite{MitrokhinCNN2020}), and the transformation into HVs was done by randomly repeating these binary vectors to get the necessary dimensionality.
To address the latter one, in~\cite{KarunaratneHDAugmented2021, KleykoAugmented2022,HerscheContinualLearn2022,herscheNVSA2022}, the authors guided a convolutional neural network to produce HDC/VSA-conforming vectors with the aid of proper attention, sharpening, and loss functions. 
The sign of produced HVs can be used to transform them into bipolar HVs (of the same dimensionality). These approaches train neural networks from scratch (as with meta-learning in~\cite{KarunaratneHDAugmented2021, KleykoAugmented2022, HerscheContinualLearn2022}, or additive loss in~\cite{herscheNVSA2022}) such that the activations of the network resemble quasi-orthogonal HVs for, e.g., images of unrelated classes. 
In~\cite{MitrokhinCNN2020, NeubertAggregation2021}, the authors superimposed HVs obtained from several neural networks, which improved the results in applications. 
Yet another promising avenue is make the processes of classification and reconstruction (i.e., generation) of raw sensory data simultaneously from each neural network. One particular realization of this idea, called ``bridge networks'', was recently presented in~\cite{OlinBridgeNetworks2021}.
Finally, it is worth mentioning that a neural network does not necessarily need to produce HVs, but it can benefit from the HDC/VSA operations by improving its retrieval performance through superimposing multiple permuted versions of an output vector, as demonstrated in \cite{DanihelkaAssociative2016}.

\subsubsection{HDC/VSA for simplifying neural networks}

In~\cite{AndersonHDDNN}, it was shown that it is possible to treat the functionality of binarized neural networks with the ideas from high-dimensional geometry. The paper has demonstrated that binarized networks work because of the properties of binary high-dimensional spaces, i.e., the properties used in Binary Spatter Codes~\cite{KanervaHyperdimensional2009}. While it is an interesting qualitative result, it did not provide a concrete way to make the two benefit from each other. 
This is not obvious, since in the standard neural networks all weights are trained via backpropagation, which is rather different from the HDC/VSA principles. 

There is, however, a family of randomized neural networks~\cite{RCNNSsurvey} where a part of the network is initialized randomly and stays fixed. There are two versions of such networks: feed-forward (e.g., random vector functional link networks~\cite{IgelnikRVFL1995} aka extreme learning machines~\cite{HuangELM2006}) and recurrent (e.g., echo state networks~\cite{ESN02} aka reservoir computing\cite{LukoseviciusRC2009}). The way the randomness is used in these networks can be expressed in terms of HDC/VSA operations for the both feed-forward \cite{KleykoDensityEncoding2020} and recurrent \cite{KleykointESN2020} versions.
Conventionally, randomized neural networks were used with real-valued representations. However, since it was realized that these networks can be interpreted in terms of HDC/VSA, it appeared natural to use binary/integer variants (as in Binary Spatter Codes and Multiply-Add-Permute) to produce activations of hidden layers of the networks. This opened the avenue for efficient hardware implementations of such randomized neural networks. 
Yet another connection between HDC/VSA and feed-forward randomized neural networks was demonstrated in~\cite{PlateAssociative2000} where it was shown that HRR's binding operation can be approximated  by such networks. 
Finally, in~\cite{BrickenAttentionSDM2021} it was shown that the address mechanism from the Sparse Distributed Memory~\cite{KanervaSDM1988} approximates the attention mechanism~\cite{VaswaniAttention2017} used in modern neural networks.

\subsubsection{HDC/VSA for explaining neural networks}

It was discussed in Section~\ref{PartI-sec:capacity}  in~\cite{KleykoSurveyVSA2021Part1} that the capacity theory~\cite{FradyCapacity2018} applies to different HDC/VSA models. As mentioned in the previous section, randomized recurrent neural networks, known as reservoir computing/echo state networks, can be formulated using HDC/VSA. Therefore, capacity theory can also be used to explain memory characteristics of reservoir computing. Moreover, using the abstract idea of dissecting a network into mapping and classifier parts~\cite{papyan2020prevalence}, it is possible to apply capacity theory for predicting the accuracy of other types of neural networks (such as deep convolutional neural networks)~\cite{KleykoPerceptron2020}.

In \cite{McCoyRNNsTensor2019}, it was shown that Tensor Product Representations approximate representations of structures learned by recurrent neural networks.

\subsubsection{The use of HDC/VSA with spiking neural networks}

Another direction of the interplay between HDC/VSA and neural networks is their usage in the context of spiking neural networks (SNN).  It is especially important in the context of emerging neuromorphic platforms~\cite{TrueNorth14,DaviesLOIHI2018}. The main advantage HDC/VSA can bring into the SNN domain is the ease of transformation to spiking activities, either with rate-based coding for HDC/VSA models with scalar components or phase-to-timing coding for HDC/VSA models with phasor components.
The Spaun cognitive architecture overviewed in Section~\ref{sec:cog:spaun} is one of the first examples where the HRR model was used in the context of SNN.
The latest developments~\cite{FradyTPAM2019, FradyKNN2020} use FHRR and HRR to implement associative memory and $k$-Nearest Neighbor classifiers on SNN. 
Further, these memories were proposed as building blocks for the realization of a holistic HDC/VSA-based unsupervised learning pipeline on SNN~\cite{OsipovHyperSeed2021}.   
While in~\cite{BentSpike2022} the Sparse Block Codes model was mapped to an SNN circuit.  
In other related efforts, an event-based dynamic vision sensor~\cite{MitrokhinSensorimotor2019,HerscheDVSCDT2020} or an SNN~\cite{ZouMemorySpiking2022, MorrisHyperSpike2022} was used to perform the initial processing of the input signals that were then transformed to HVs to form the prediction model.

These works provide some initial evidence of the expressiveness of HDC/VSA on the one hand, and compatibility with SNNs on the other.  
We therefore foresee  that using HDC/VSA as a programming/design abstraction for various cognitive functionalities will soon manifest itself in the emergence of novel SNN-based applications.   

\subsubsection{``Hybrid'' solutions}
\label{sec:disc:nn:hybrid}

By ``hybrid'' in this context we refer to solutions that use both neural networks and some elements of HDC/VSA. 
Currently, a particularly common primitive used in such hybrid solutions is the representation of a set of role-filler bindings, or superposition of multiplicative bindings.  
For example, in~\cite{CheungSuperposition2019} the weights of several neural networks were stored jointly by using the superposition operation, which alleviated the problem of ``catastrophic forgetting''.  
In~\cite{WilsonOOD2021}, activations of layers of a deep neural network were used as filler HVs.
They were bound to the corresponding random role HVs and all role-filler bindings were aggregated in a single superposition HV that in turn was used to successfully detect out-of-distribution data.
Similarly, in~\cite{NeubertAggregation2021, NeubertPlaceRecognition2021,SutorGluing2022}, activations of several neural networks were combined together via HDC/VSA operations. 
In~\cite{NeubertAggregation2021, NeubertPlaceRecognition2021}, this idea was used to form a single HV compactly representing the aggregated neural networks-based image descriptor while in~\cite{SutorGluing2022} outputs of multiple neural networks were fused together to solve classification problems.
In~\cite{GanesanLearning2021}, the superposition of role-filler bindings was used to simultaneously represent the output of a deep neural network when solving multi-label classification tasks.
In~\cite{herscheNVSA2022}, the activations of the last layer generate a query HV that resembles the superposition of the visual objects available in a panel, whereby each object is uniquely represented by multiplicative binding of its attributes' HVs.
In addition, a review of hybrid solutions combining Tensor Product Representations and neural networks such as Tensor Product Generation Networks~\cite{HuangTensorNetworks2018} can be found in~\cite{SmolenskyNeurocompositional2022}.
Finally, it is worth noting that all these solutions in some way 
relied on the idea of ``computing in superposition''~\cite{KleykoComputingParadigm2021} suggesting that HVs can be used to simultaneously manipulate several pieces of information.

\subsection{Open issues}

As introduced at the beginning of this survey, HDC/VSA originated from proposals of integrating the advantages of the symbolic approach to Artificial Intelligence, such as compositionality and systematicity, and those of the neural networks approach to Artificial Intelligence (connectionism), such as vector-based representations, learning, and grounding. 
There is also the ``neural-symbolic computing'' ~\cite{Neural-symbolic_Book_2002,Garcez2019NSC2019}, or ``neurosymbolic AI''~\cite{GarcezNeurosymbolic2020} community that suggests hybrid approaches to Artificial Intelligence. The key idea is to form an interface so that symbolic and connectionist approaches can work together. 
At present, HDC/VSA and neural-symbolic computing seem to be rather separate fields that can benefit from synergy.
Moreover, the works developing cognitive architectures and Artificial Intelligence with HDC/VSA are rather limited~\cite{EliasmithSPAUN2012, RachkovskijWorldModel2013, EliasmithBuildBrain2013}.

So far, a major advantage of the HDC/VSA models has been their ability to use HVs in a single unified format to represent data of varied types and modalities. Moreover, the use of HDC/VSA operations allows introducing compositionality into representations. The prerequisite, however, is that the representation of the input data to be transformed into HVs should be able to specify the compositionality explicitly. Nevertheless, despite this advantage, HDC/VSA is usually overlooked in the context of neural-symbolic computing, which calls for establishing a closer interaction between these two communities.

In fact, most of the works use HDC/VSA to reimplement the symbolic Artificial Intelligence primitives or the machine learning functionality with HVs, in a manner suitable for emerging unconventional computing hardware. 
At the same time, when transforming data into HVs, machine learning is used rarely, if at all. Learning is used mainly for training a classifier based on the already constructed HVs.
In most recent studies, the narrative is often to demonstrate solutions to some simple classification or similarity search problems. In so, the quality of the results is comparable to the state-of-the-art solutions, but the energy/computational costs required by an HDC/VSA-based solution are only a fraction of those of the baseline approaches. 
These developments suggest that HDC/VSA might find one of their niches in application areas known as ``tiny machine learning'' and ``edge machine learning''. 
Nevertheless, there is an understanding that manually designing transformations of data into HVs for certain modalities (e.g., 2D images) is a challenging task. This stimulates attempts that use modern learning-based approaches, such as deep neural networks, for producing HVs (see Section~\ref{disc:ANNs:produce}). The current attempts, however, are rather limited since they focus heavily on using HVs formed from the representations produced by neural networks to solve some downstream machine learning tasks (e.g., similarity search or classification).

Another approach would be to discover general principles for combining neural networks and HDC/VSA, but currently there are only few such efforts~\cite{CheungSuperposition2019, ChenTPRMapping2020, HerscheCompressingBCI2020,  OlinBridgeNetworks2021, ZemanCompressed2021,herscheNVSA2022} (see also Section~\ref{sec:disc:nn:hybrid}). For example, the Tensor Product Representations operations in~\cite{ChenTPRMapping2020}, and the HDC/VSA operations in~\cite{herscheNVSA2022}, are introduced into the neural network machinery. 
These attempts are timely for the connectionist approach to Artificial Intelligence since, despite recent uptick of deep neural networks~\cite{LeCunDL2015}, there is a growing awareness within the connectionist community that reaching Artificial General Intelligence is going to require a much higher level of generalization than that available in modern neural networks~\cite{MarcusAI2020, GreffBinding2020, GoyalDLCognition2020, HintonPartwhole2021}. 
A recent proposal in~\cite{SmolenskyNeurocompositional2022} could be considered more HDC/VSA-oriented.

One of the milestones toward reaching Artificial General Intelligence is solving the problem of compositionality. For example,~\cite{GreffBinding2020} stressed the importance of various aspects of binding for achieving compositionality and generalization. The framework of HDC/VSA has a dedicated on-the-fly operation for binding (Section~\ref{PartI-sec:binding} in~\cite{KleykoSurveyVSA2021Part1}), which does not require any training. The neural implementation of binding in the context of SNNs is still an open issue. There are, however, two recent proposals~\cite{RennerBinding2022, BentSpike2022} aiming at addressing this issue.

Further, \cite{GoyalDLCognition2020} suggested that various inductive biases are required for human-level generalization, including compositionality and discovery of causal dependencies. HDC/VSA have a potential to achieve this through analogical reasoning  (Section~\ref{sec:analogical:reasoning}). However, the progress is held back by the lack of mechanisms to build the analogical episodes, e.g., by observing the outer world or by just reading texts. The analogical episodes should include two major types of hierarchy, the compositional (``part-whole'') one, and the generalization (``is-a'') hierarchy. We believe that associative memories~\cite{GritsenkoAMSurvey2017} may provide one way to form ``is-a'' hierarchies (see the discussion in~\cite{RachkovskijWorldModel2013}), but this topic has not yet been studied extensively in the context of HDC/VSA. In terms of forming part-whole hierarchies from 2D images, a recent conceptual proposal was given in~\cite{HintonPartwhole2021}. The essence of the proposal is learning to parse 2D images by training a neural network and using the similarity of the learnt high dimensional vector representations for analogical reasoning. 
An interesting direction for future work is to see how such representations can be paired with the analogical reasoning capabilities of HDC/VSA.
However, all the above and other proposals from the connectionist community rely on learning all the functional lacking in neural networks. But there are also discussions on the innate structure of neural networks~\cite{MarcusAI2020}. For the sake of fairness, it should be noted that current HDC/VSA also lack ready implementations for most of the mentioned above functionality.
There are a lot of open problems that should be addressed to build truly intelligent systems. Below we list some of them.
Some problems related to the internals of HDC/VSA are the following:
\begin{itemize}

\item Recovery. Recovering element HVs from compositional HVs. Section~\ref{PartI-sec:recovery} in~\cite{KleykoSurveyVSA2021Part1} presented some ideas for recovering the content of HVs. However, for most of the HDC/VSA models, knowledge of all but one bound HV is required. This makes the recovery problem combinatorial (but see a proposal in~\cite{FradyResonator2020, KentResonatorNetworks2020}).

\item Similarity. In many of the HDC/VSA models, the results of the binding operation are not similar as soon as a single input HV is dissimilar. 
While it is often convenient, or even desired, that the result of the binding operation is dissimilar to its input HVs, the price to be paid is weak or no similarity in the resultant HV that includes the same HVs in slightly different combinations. This might hinder the similarity search that is at the heart of HDC/VSA and required in many operations such as the recovery or clean-up procedures. 

\item Memory. Quantity and allocation of item memories. How many item memories and which HVs of all available HVs should be placed in each of the? Types of the item memories to be used? List memories provide reliable retrieval, but are problematic for generalization. Distributed auto-associative memories have problems with dense HVs (but see~\cite{RamsauerHopfield2020}).

\item Generalization of HVs. How to form generalized HVs containing typical features of objects and, is-a hierarchies of objects, but preserve HVs of specific objects as well? Distributed auto-associative memories have potential for generalization by unsupervised learning~\cite{GritsenkoAMSurvey2017}. 

\item Generativity. Is it possible to make a new meaningful compositional HV without constructing it from atomic HVs? Is it possible to produce meaningful input (e.g., fake image or sound, as in deep neural networks) from some generated generalized or specific compositional HV?

\item Similarity-based associations. How to select the particular association needed in the context from myriads of possible associations? For example, between HVs of a part and of a whole, or between HVs of a class and of a class instance. 

\item Parsing. How to parse the raw sensory inputs into a meaningful part-whole hierarchy of objects?

\item Holistic representation. HVs are holistic representations. However, for the comparison of objects we may need to operate with their parts. Part HVs can be found given a particular holistic HV. Is it possible to form holistic HVs of very different inputs of the same class so that they are similar enough to be found in memory?

\item Dynamics. Representation, storage, similarity-based search, and replaying of spatial-temporal data (e.g., a kind of video).

\item Probabilistic representation. Representation of object(s) reflecting the probabilities assigned to them. 

\item Learning. How to learn, for example, the most suitable transformation of input data into HVs for a given problem? Also, learning HV for behaviors, including reinforcement learning. 
\end{itemize}

Let us also touch on problems specific not  only to HDC/VSA:

\begin{itemize}
\item Representation invariance. To recognize the same object in various situations and contexts, we need some useful invariance of representation that makes similar diverse manifestations of the same object. 

\item Context-dependent representation. Representing data as objects and getting the proper representation of an object in a particular context. 

\item Context-dependent similarity. For example, depending on the context a pizza is similar to a cake but also to a frisbee. How can such a context-dependent similarity be implemented? 

\item Context-dependent processing. The type of processing to be applied to data should take into account the context, such as bottom-up and top-down expectations or system goals.

\item Hierarchies. Forming ``part-whole'' and ``is-a'' hierarchies. Which level of ``part-whole'' hierarchy does an object belong to? An object can belong to various levels for different scenes of the same nature. This is also connected to the image scale. An object can belong to very many hierarchies for scenes of varied nature. Concerning ``is-a'' (class-instance) hierarchy, an object can belong to various hierarchies of classes, subclasses, etc. in different contexts. 

\item Cause-effect. Cause-effect extraction in new situations could be done by analogy to familiar ones. Generalizations and specifications using is-a hierarchy are possible. 

\item Interface with symbolic representations. 
Designers of cognitive agents have to solve the dual problem of both building a world model and building it so that it can be expressed in symbols in order to interact with humans. 

\item The whole system control. Most of the solutions generally rely on conventional deterministic mechanisms for flow control. It is, however, likely that the control of the system should also be trainable, so that the system could adjust it for new tasks. 
\end{itemize}

All the problems described above are rarely (if at all) considered in the scope of HDC/VSA. To the best of our knowledge, one study that discussed some of these problems is~\cite{RachkovskijWorldModel2013}. 
Moreover, it is not fully clear which of these problems are related to the general problems of building Artificial General Intelligence, and which ones are due to the architectural peculiarities of neural networks and HDC/VSA. In other words, the separation provided above is not necessarily unequivocal. We believe, however, that building Artificial General Intelligence will require facing these problems anyway. 
Finally, we hope that insights from HDC/VSA, symbolic Artificial Intelligence, and neural networks will contribute to the solution.

\section{Conclusion}
\label{sec:conc}

In this two-part survey, we provided comprehensive coverage of the computing framework known under the names Hyperdimensional Computing and Vector Symbolic Architectures.
Part~I of the survey~\cite{KleykoSurveyVSA2021Part1} covered existing models and transformations of input data of various types into distributed representations.
In this Part~II, we focused on known applications of Hyperdimensional Computing/Vector Symbolic Architectures including the use in cognitive modeling and cognitive architectures.
We also discussed the open problems along with  promising directions for the future work. 
We hope that for newcomers, this two-part survey will provide a useful guide of the field, as well as facilitate its exploration and the identification of fruitful directions for research and exploitation. 
For the practitioners, we hope that the survey will broaden the vision of the field beyond their specialization. 
Finally, we expect that it will accelerate the convergence of this interdisciplinary field to discipline with common terminology and solid theoretical foundations.

\bibliographystyle{IEEEtran} 
\bibliography{Bibliography}

\end{document}